\documentclass[12pt]{article}

\pdfoutput=1
\usepackage{subfigure}
\usepackage{amsmath, amssymb, amsfonts}
\usepackage{bm}
\usepackage{makecell}
\usepackage[normalem]{ulem}
\usepackage{booktabs}
\usepackage{multirow}
\usepackage{hyperref}
\usepackage{geometry}
\usepackage{authblk}
\usepackage{marvosym}
\usepackage{graphicx}

\usepackage{times}
\usepackage{mathptmx}
\hypersetup{ 
    colorlinks=true, 
    citecolor=[rgb]{0.2,0.5,0.9}, 
    linkcolor=[rgb]{0.2,0.5,0.9},
    anchorcolor=[rgb]{0.2,0.5,0.9},
    urlcolor=[rgb]{0.2,0.5,0.9}
    }
\providecommand{\keywords}[1]
{
  \textbf{\text{Keywords: }} #1
}

\begin{document}
\title{Machine Learning for Infectious Disease Risk Prediction: A Survey}

\author[a]{Mutong LIU $^{\href{mailto:csmtliu@comp.hkbu.edu.hk}{\textrm{\Letter}}}$}
\author[a]{Yang LIU \thanks{Corresponding author: csygliu@comp.hkbu.edu.hk}$^{\href{mailto:csygliu@comp.hkbu.edu.hk}{\textrm{\Letter}}}$}
\author[a]{Jiming LIU $^{\href{mailto:jiming@comp.hkbu.edu.hk}{\textrm{\Letter}}}$}
\affil[a]{Department of Computer Science, Hong Kong Baptist University, Hong Kong SAR, China}

\date{\vspace*{-1cm}}
\maketitle

\begin{abstract}
Infectious diseases, either emerging or long-lasting, place numerous people at risk and bring heavy public health burdens worldwide. In the process against infectious diseases, predicting the epidemic risk by modeling the disease transmission plays an essential role in assisting with preventing and controlling disease transmission in a more effective way.
In this paper, we systematically describe how machine learning can play an essential role in quantitatively characterizing disease transmission patterns and accurately predicting infectious disease risks.
First, we introduce the background and motivation of using machine learning for infectious disease risk prediction. Next, we describe the development and components of various machine learning models for infectious disease risk prediction. Specifically, existing models fall into three categories: Statistical prediction, data-driven machine learning, and epidemiology-inspired machine learning. Subsequently, we discuss challenges encountered when dealing with model inputs, designing task-oriented objectives, and conducting performance evaluation. Finally, we conclude with a discussion of open questions and future directions.
\end{abstract}

\newpage
\tableofcontents
\newpage

\section{Introduction}
% \textbf{{\color{red}(Use covid-19 and malaria as two examples will be good. Show how serious they are by data or statistics.)
The propagation of infectious diseases, whether emergent (e.g., coronavirus disease 2019 (COVID-19), which is responsible for the ongoing pandemic and has caused nearly 7 million deaths worldwide so far \footnote{https://covid19.who.int/. Accessed April 29, 2023}) or long-standing (e.g., malaria, which has an ancient history and still causes more than 600 thousand deaths every year \cite{world2022world}), significantly affects human well-being and social development on a global scale \cite{world2022world, chakraborty2020covid}.
% (e.g., malaria and dengue with an ancient history \cite{gubler1998dengue})
%For example, according to the world malaria report 2020, estimated 229 million malaria cases and about half a million death people occurred in 2019 in 87 malaria-endemic countries~\cite{world2020world}.
Thus, the battle against infectious disease is never-ending. Humankind's development of countermeasures to various diseases has been based on conceptual innovation together with scientific development in multiple disciplines, from the use of vaccination to eradicate smallpox, a high-mortality disease, to the use of combinations of multiple interventions to contain the transmission of severe acute respiratory syndrome coronavirus 2 (SARS-CoV-2), which causes COVID-19.
% such as clinical medicine, epidemiology, and data science
% and collaboration of multiple institutions and countries
% these achieved great success for humankind, such as the eradication of smallpox by vaccination.
% Along with the, humankind has achieved great success in these battles.

% \textbf{{\color{blue}(Introduce the focuses of prediction on different phases of disease transmission)}}
% {\color{red}See if it is necessary to put a paragraph here to smoothly narrow down the focus of the paper from the whole infectious disease area to data-science related topics.}

In recent decades, machine learning has been widely and successfully applied in many fields, e.g., natural language processing and computer vision, to perform various tasks, e.g., regression and classification. Inevitably, due to the ability of machine learning to deal with large and heterogeneous data and capture complex patterns, it has also been employed in infectious disease research \cite{agrebi2020use, santangelo2023machine}.
% as well as the prediction of infectious disease outbreaks.
% In the recent decade, data science, which is promoted by rich data and a surge of big data-based studies, has taken its role in many domains. Although it may not provide explanations as satisfying as the traditional experiments in some areas, such as clinical medicine and biology, it undoubtedly provides a promise to discover unknowns in a more effective way by constructing prediction models \cite{agarwal2014big}. Inevitably, the development of data science also swept the field of predictive epidemic risk modeling \cite{y2019charting}.
However, the focus of coping with disease risks varies across the different stages of infectious disease progression, due to the different goals of public health strategies (e.g., the prevention, mitigation, or containment of disease transmission), and thus the goal of disease modeling for prediction of disease risk also varies across these stages. With reference to \cite{han2016future}, the developmental process of infectious disease risk can be divided into three phases: the watch phase, the warning phase, and the emergency phase.
In the watch phase, an infectious disease has not yet occurred in humans but possibly exists in the environment surrounding the areas where humans live. Thus, in this phase, data-driven modeling is used to investigate hosts (e.g., wild animals) that may carry pathogens (e.g., microorganisms that could cause infectious diseases, such as certain species of bacteria, viruses, protozoa, fungi, and prions), as this is essential to prevent transmission of pathogens from hosts to humans and thereby prevent outbreaks in human populations.
If these pathogens come into contact with humans under specific conditions, human infection can occur; thus, once human cases have been verified, the level of infectious disease risk is upgraded to the warning phase. In this phase, it is crucial for disease transmission to be understood to allow public health agencies to appropriately respond. The importance of applying phylogenetic and phylodynamic modeling techniques in this phase was highlighted, as these techniques can help us to understand the properties and potential of disease transmission by offering informative predictions in situations in which spatiotemporal disease transmission data are scarce \cite{attwood2022phylogenetic}.
If an epidemic is not efficiently brought under control or the efforts of interventions are overwhelmed due to the rapid spread of a disease, the epidemic can spread over such a large geographical range that it becomes a pandemic, leading to high morbidity and mortality, such as the COVID-19 pandemic. In such situations, the level of infectious risk enters the emergency phase. As a result, data-informed modeling and prediction of disease risk and severity (such as infection size, scope, and duration), which are recognized as the main focus in the influenza season Challenges held by the Centers for Disease Control (CDC) in the United States \cite{biggerstaff2016results}, and exploration of the effect of available intervention strategies \cite{prem2020effect}, are more urgent in the emergency phase than in previous phases, as such investigations are needed to inform decision-makers on how to take action to achieve the goal of reducing damage to human health.

% \textbf{{\color{blue}(Emphasis the focus of our paper and introduce the importance of epidemic prediction)}}
In this paper, we focus on machine learning approaches for infectious disease risk prediction in the emergency phase. Disease risk prediction can provide insights and quantitative information for decision-making processes on how to contain disease transmission and mitigate loss, so it is considered an important aid for the formulation of public health responses \cite{lutz2019applying}.
Specifically, accurate prediction of epidemic or pandemic trends can provide advance warning of potential outbreaks and thus allow timely action to be taken, such as the allocation of anti-disease resources to regions with urgent needs or the adjustment of quarantine policies \cite{chitnis2010mathematical} to prevent an outbreak as quickly as possible. Furthermore, retrospective analysis of disease trends based on prediction models can reveal the transmission patterns underlying observations and enable future outbreaks to be dealt with more effectively than the current outbreak \cite{shaman2012forecasting}.
Given that big data technologies and machine learning models have been successfully applied in many fields, increasing numbers of researchers in the fields of machine learning and statistics are considering how to utilize large-scale available data and the capacity of machine learning for data representation and data fitting to accurately predict disease risks. Numerous novel infectious disease risk prediction models have been devised, with a range of goals. Initially, the main focus of most studies was to develop statistical prediction models and data-driven machine learning models by designing various model structures to automatically capture implicit dependencies based on observed data and by minimizing prediction errors.
However, when such models are used in practice, accurate prediction based on statistical relationships is not the ultimate goal, and questions continually emerge. For example, how do we know we can trust the predictions of a data-driven model? What information can be provided by predictions? These questions reflect the importance of accurate, informative, and interpretable predictions, as they provide reliable and valid information for disease prevention and control. In recent studies, this has been achieved by integrating prior knowledge of epidemiological models with data-driven models to create epidemiology-inspired machine learning models.

%\textbf{{\color{blue}(Introduce previous surveys and their focuses)}}
Over the past two decades, several authors have summarized progress in the development of infectious disease models. For instance, Grassly and Fraser \cite{grassly2008mathematical} examined multiple topics on the linking of mathematical hypotheses and modeling to the process of infectious disease transmission, and summarized studies that have devised mathematical models of infectious disease transmission. In recent years, some surveys have been performed on models for specific infectious diseases, such as malaria \cite{mandal2011mathematical}, dengue \cite{andraud2012dynamic}, influenza \cite{nsoesie2014systematic, chretien2014influenza}, and COVID-19 \cite{adiga2020mathematical, clement2021survey}. \cite{mandal2011mathematical} and \cite{andraud2012dynamic} have introduced and summarized the development of determinate or stochastic mathematical approaches for modeling malaria and dengue transmission, and the evolution of elaborate hypotheses on these processes. These surveys have covered several areas, such as population-level compartmental models (also known as mass-action compartmental models), structured metapopulation models, and agent-based models, but have not paid much attention to machine learning models, which have undergone continual development over the past decade.
% statistical models and hybrid models, such as
\cite{nsoesie2014systematic} and \cite{chretien2014influenza} have surveyed studies on mechanism-based models and data-driven models for influenza forecasting. The two most recent published surveys are those of \cite{adiga2020mathematical} and \cite{clement2021survey} and these have examined many state-of-the-art deep learning models. For example, \cite{adiga2020mathematical} sorted mathematical models of COVID-19 into three categories: statistical, mechanistic, and hybrid models. In contrast, \cite{clement2021survey} sorted computational models of COVID-19 transmission and diagnosis more finely, to give five categories: compartmental, statistical, data-driven, machine learning- and deep learning-based, and mixed models. 
However, although the categories in \cite{adiga2020mathematical} and \cite{clement2021survey} encompass epidemiologically inspired models, these surveys mainly focused on mechanism-based models and statistical models, and therefore did not cover some recently developed epidemiologically inspired machine learning models.

In this paper, we categorize, summarize, and discuss machine learning methods for infectious disease risk prediction. To this end, we first provide an overview of the previously developed machine learning models by sorting them into three categories: (1) statistical prediction models (Section~\ref{Statistical prediction}), (2) data-driven machine learning models (Section~\ref{Data-driven machine learning}), and (3) epidemiology-inspired machine learning models (Section~\ref{Epidemiology-inspired machine learning}).
Next, we briefly introduce a selection of related methods in each category to show their innovations, differences, and similarities. Subsequently, we discuss in three sections the three kinds of challenges commonly faced when predicting epidemic risk: (1) data-related challenges (Section~\ref{Data challenges}); (2) task-related challenges (Section~\ref{Task challenges}); and (3) evaluation-related challenges (Section~\ref{Evaluation challenges}). In each section, we cover a series of sub-topics and introduce the techniques that have typically been used to address these challenges.

\section{Machine learning for infectious disease risk prediction}
As mentioned, we divide current machine learning approaches for infectious disease risk prediction into three categories. The first category is statistical prediction models and contains studies that have mostly treated epidemic prediction as a time-series prediction problem to analyze the statistical characteristics of trends; the second category is data-driven machine learning models and contains many studies that have presumed that there are implicit and unknown disease propagation patterns (such as spatiotemporal transmission networks) that can be captured by various model structures, whose parameters can be inferred from available data; and the third category is epidemiology-inspired machine learning models and comprises studies that have incorporated prior knowledge of epidemiological models with the inferential ability of data-driven machine learning to better characterize disease transmission than previous approaches.

\begin{figure*}[htb!]
% \vspace{0.1cm}
  \centering
  \subfigure{\includegraphics[width=1\textwidth]{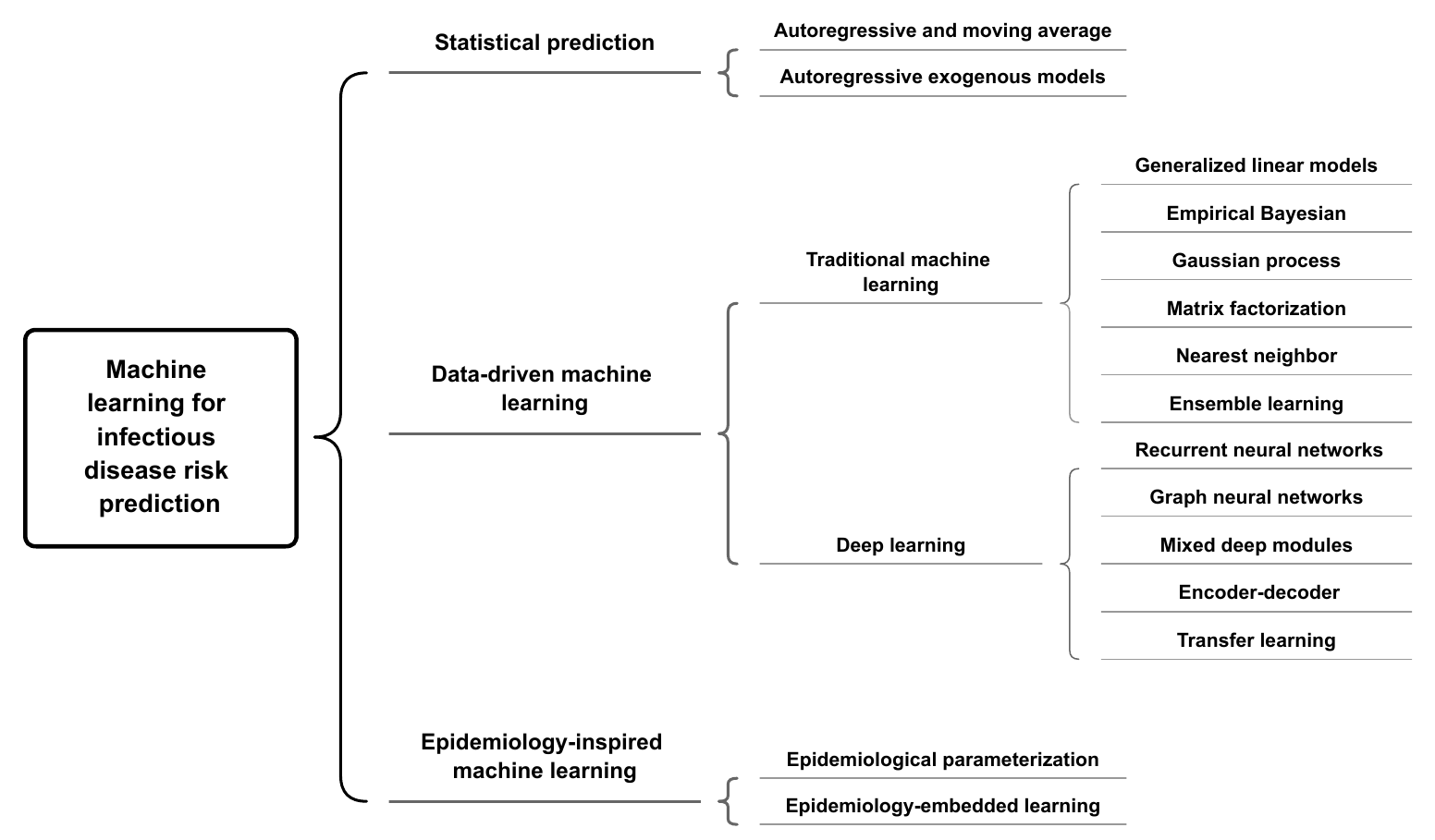}}
  \caption{Categories of machine learning models that have been developed for infectious disease risk prediction: (1) Statistical prediction models (Section~\ref{Statistical prediction}); (2) data-driven machine learning models (Section~\ref{Data-driven machine learning}); and (3) epidemiology-inspired machine learning models (Section~\ref{Epidemiology-inspired machine learning}).}
  \label{fig:structure of models}
 % \vspace{0.1cm}
\end{figure*}

% Based on different scenarios, various models for modeling the disease transmission network have been developed.
% Specifically, works of statistical prediction and data-driven machine learning usually do not define explicit transmission mechanisms to mathematically depict the disease transmission but rather set or learn the model structure and parameters based on the observed data.
% Different from mechanism-based models, data-driven models for epidemic prediction
In this section, we provide a brief introduction to the models in each category and then subdivide these categories to show their differences and relationships. The specific structure of the taxonomy we employ to classify the models is given in Fig.~\ref{fig:structure of models}.

\subsection{Problem statement}
% over a time period in one location or multiple locations
In the task of predicting infectious disease risks, given observations of disease dynamics-related data $\boldsymbol{x}$, the goal of machine learning approaches is to train a model $f$ to accurately predict the future disease dynamics $y$ in one location or multiple locations using historical data:
\begin{equation}
    y=f(\boldsymbol{x}).
\end{equation}
Usually, $\boldsymbol{x}$ denotes the model input, and it could be the historical data of the indicator of disease risks, such as the disease case number or disease prevalence. In some circumstances, it could also include other risk-related data, such as climate data, mobility data, and population data, to enhance the prediction performance. $y$ is the model output, i.e., the future risks to be predicted, which is usually the indicator of infectious disease risks. In general, in the temporal dimension, the input and output could cover one time step or multiple time steps; in the spatial dimension, the input and output could cover one location or multiple locations. In the following content, we use $\boldsymbol{x}$ to denote the model input, $y$ to denote the model output, and we do not make assumptions on their dimensions. We use $f$ to denote the general format of model's prediction functions; the specific formulations of them depend on the used model structures in different works.
% is usually sequenced over time, sometimes for one location, and sometimes for multiple locations.
% $\boldsymbol{D}=\{\boldsymbol{D}_1,\boldsymbol{D}_2,...,\boldsymbol{D}_T\}^\mathrm{T} \in \mathbb{R}^{T\times N}$, where $N$ denotes the number of locations and $T$ denotes the time range, is often given.
% For each time step $t$, $\boldsymbol{D}_t=\{D_{t,1},D_{t,2},...,D_{t,N}\}^\mathrm{T} \in \mathbb{R}^{N\times 1}$, where $D_{t,i}$ corresponds to the disease dynamics (such as case number) of location $i$ at time $t$.

\subsection{Statistical prediction}\label{Statistical prediction}
Because epidemic or pandemic data are usually presented in time-series form, the task of infectious disease risk prediction can usually be treated as a time-series prediction problem. Therefore, many statistical models are applied to epidemic or pandemic prediction.
Some classic statistical models are based on linear model structures, such as \textbf{a}utore\textbf{g}ressive (AR) models, which linearly combine past observations within a time window $p$ (i.e., $x_{t-1}$, $x_{t-2}$, $\cdots$, $x_{t-p}$) with the disturbance term $E_t$ (the AR model equation is shown in Eq.~\ref{Eq:AR} \cite{dettling2013applied}); \textbf{m}oving \textbf{a}verage (MA) models, which complement AR models by linearly combining disturbance terms within a time window $q$ (i.e., $E_t,E_{t-1},\cdots,E_{t-q}$) (the MA model equation is shown in Eq.~\ref{Eq:MA} \cite{dettling2013applied}); and their combination, known as autoregressive moving average (ARMA) models (the model equation is shown in Eq.~\ref{Eq:ARMA} \cite{dettling2013applied}), which characterize the features of single time-series dynamics to forecast future trends \cite{dettling2013applied, lutkepohl2004applied}.
\begin{equation}\label{Eq:AR}
y\ or\ x_{t}=\alpha_{1} x_{t-1} + \alpha_{2} x_{t-2}+ \cdots + \alpha_{p} x_{t-p} + E_{t}
\end{equation}
\begin{equation}\label{Eq:MA}
y\ or\ x_{t}=E_{t} + \beta_{1} \cdot E_{t-1}+ \cdots +\beta_{q} \cdot E_{t-q}
\end{equation}
\begin{equation}\label{Eq:ARMA}
y\ or\ x_{t}= \ \alpha_{1} x_{t-1}+\cdots+\alpha_{p} x_{t-p} + E_{t}+\beta_{1} E_{t-1}+\cdots+\beta_{q} E_{t-q}
\end{equation}

However, the above-mentioned models can only be applied to time series that are stationary, and so cannot be applied in many situations, as time series are often non-stationary due to the effects of seasonal factors, persistent interventions, or other determining factors. Therefore, variations of the above-mentioned models have been proposed to cope with these situations.
For example, \textbf{a}utore\textbf{g}ressive \textbf{i}ntegrated \textbf{m}oving \textbf{a}verage (ARIMA) models \cite{dettling2013applied} remove obvious trends (such as upward or downward trends)---which are caused by determining factors---by using $d$-order differencing processes. This affords stationary time series, to which   ARMA models can be applied. \textbf{S}easonal ARIMA models \cite{dettling2013applied} remove the effects of seasonality by performing lagged differencing processes with a period $s$. \textbf{A}uto\textbf{r}egressive e\textbf{x}ogenous (ARX) models also use disease dynamics data to make predictions but also take other risk-related factors into account; these are denoted extra or exogenous variables and formulated as a weighted sum item that is in addition to the original weighted sum item.
Wang et al. developed a variation of the standard ARX model that they denoted the \textbf{d}ynamic \textbf{P}oisson \textbf{a}utoreg\textbf{r}essive model with e\textbf{x}ogenous inputs variables (DPARX) model, whose parameters dynamically change over time \cite{wang2015dynamic}. That is, the DPARX model learns a set of parameters for a prediction at each time step, which results in many parameters to learn. To effectively do so and avoid overfitting, Wang et al. assumed that some models at different time steps---and thus also their parameters---are similar to each other. They provided three types of prior constraints with graphical structures (i.e., fully connected, nearest neighbors connected, and seasonal nearest neighbors connected structures) to depict this kind of structural similarity between models at different time steps.

\subsection{Data-driven machine learning}\label{Data-driven machine learning}

\begin{table}[htbp]
  \centering
  \caption{Summary of data-driven machine learning models.}
  \label{tab:Data-driven machine learning}%
    % \begin{tabular}{c|c|p{6em}p{15.75em}}
    \scalebox{0.96}{
    \begin{tabular}
    {m{2.5cm}<{\centering}|m{2cm}<{\centering}|cm{8.5cm}<{\centering}}
    \toprule
    \multicolumn{2}{m{4cm}<{\centering}}{\textbf{Categories}} & \textbf{References} & \textbf{Data-driven components} \\
    \midrule
    \multirow{8}[26]{2cm}{\centering Traditional machine learning} & \multirow{2}[6]{2cm}{\centering Generalized linear models} & \cite{zhang2015unified} & Poisson regression \\
    \cmidrule{3-4} & & \cite{pei2018group, pei2020active} & Gaussian regression for continuous variables, and Bernoulli regression for discrete variables \\
    \cmidrule{2-4} & Empirical Bayesian & \cite{brooks2015flexible} & Empirical Bayes framework \\
    \cmidrule{2-4} & \multirow{2}{2cm}{\centering Gaussian process} & \cite{senanayake2016predicting} & Gaussian process with spatial and temporal kernel for spatiotemporal dependency \\
    \cmidrule{3-4}  &  & \cite{zimmer2020influenza} & Gaussian processes for temporal dependency \\
    \cmidrule{2-4} & Matrix factorization and nearest neighbor & \cite{chakraborty2014forecasting} & Matrix factorization based regression using nearest neighbor embedding \\
    \cmidrule{2-4} & \multirow{2}[4]{2cm}{\centering Ensemble learning} & \cite{chakraborty2014forecasting} & Fusion of multiple models trained by different data sources \\
    \cmidrule{3-4} &  & \cite{zimmer2020influenza} & Fusion of multiple regression models trained by different features of data \\
    \midrule
    \multirow{13}[26]{*}{Deep learning} & \multirow{3}[6]{*}{RNN} & \cite{tan2020demystifying} & Hierarchically stacked RNN \\
    \cmidrule{3-4}   &  & \cite{venna2018novel}  & Sequentially stacked LSTM \\
    \cmidrule{3-4}   &  & \cite{volkova2017forecasting} & Independent two LSTM layers and a fusion layer \\
    \cmidrule{2-4}   & \multirow{2}[4]{*}{GNN} & \cite{kapoor2020examining} & GNNs with spatial and temporal connections \\
    \cmidrule{3-4}   &  & \cite{cao2020spectral} & GCN combining with GFT and DFT \\
    \cmidrule{2-4}   & \multirow{5}[10]{2cm}{\centering Mixed deep modules} & \cite{wu2018deep} & CNN and RNN for spatial and temporal dependency respectively \\
    \cmidrule{3-4}   &  & \cite{deng2020cola} & Dilated convolution and RNN for temporal dependency and GNN with attention matrix for spatial dependency \\
    \cmidrule{3-4}   &  & \cite{zheng2021hierst} & LSTNet/N-Beats and GAT for temporal and spatial dependency respectively \\
    \cmidrule{3-4}   &  & \cite{kamarthi2021doubt} & Functional neural process \\
    \cmidrule{3-4}   &  & \cite{gao2022popnet} & GAT for spatial dependency, GRU and dilated CNN for temporal dependency \\
    \cmidrule{2-4}   & \multirow{2}[4]{2cm}{\centering Encoder-decoder} & \cite{adhikari2019epideep} & LSTM and IDEC to encode and cluster the temporal dependency respectively\\
    \cmidrule{3-4}   &  & \cite{cui2021into} & Multi-channel CNNs to encode temporal dependency and GRE for encode spatiotemporal dependency \\
    \cmidrule{2-4}   & Transfer learning & \cite{rodriguez2021steering} & EpiDeep as source model and CAEM to encode spatiotemporal features \\
    \bottomrule
    \end{tabular}%
    }
\end{table}%

\subsubsection{Traditional machine learning models}

\ 

\noindent
In addition to a series of statistical autoregressive models that consider time dependency, many machine learning models have been developed that use more flexible model structures to predict disease risk. A summary of the data-driven machine learning models is given in Table \ref{tab:Data-driven machine learning}.

\paragraph{Generalized linear models}
Some studies have used \textbf{g}eneralized \textbf{l}inear \textbf{m}odels (GLMs) to predict disease risks in a single location or multiple locations. The general formulation of GLMs is given as follows \cite{murphy2022probabilistic}:
\begin{equation}\label{GLMs}
    p(y|\boldsymbol{x}, \boldsymbol{w}, \sigma^2) = \exp \left[ \frac{y\boldsymbol{w}^{T}\boldsymbol{x}-A(\boldsymbol{w}^{T}\boldsymbol{x})}{\sigma^2}+\log h(y, \sigma^2) \right],
    % = \exp \lbrack \frac{y\boldsymbol{w}^{T}\boldsymbol{x}-A(\boldsymbol{w}^{T}\boldsymbol{x})}{\sigma^2}+\log h(y, \sigma^2) \rbrack,
    % \eta = \boldsymbol{w}^{T}\boldsymbol{x},
\end{equation}
where $y$ denotes the response variable to be predicted, $\boldsymbol{x}$ denotes the input feature vector used to predict $y$,  $\boldsymbol{w}$ denotes the weighting parameter vector on $\boldsymbol{x}$, $\sigma^2$ represents the dispersion term, $A(\cdot)$ represents the log normalizer, $h(\cdot, \cdot)$ represents the base measure, and $\exp \lbrack \cdot \rbrack$ denotes a given probability distribution from the exponential family.
%The $h(\cdot)$, $A(\cdot)$, $\mathcal{T}(\cdot)$ are known in distributions. 
Furthermore, there is a mean function $g^{-1}$ that maps $\boldsymbol{w}^{T}\boldsymbol{x}$ to the mean value of the response variable: $\mu = g^{-1}(\boldsymbol{w}^{T}\boldsymbol{x})$~\cite{murphy2022probabilistic}.
In different instances, different probability distributions are used to model disease risks. 
For example, in \cite{zhang2015unified}, Zhang et al. used the Poisson distribution to model case numbers as integer values. 
In \cite{pei2018group, pei2020active}, Pei et al. used the Gaussian distribution to model the disease case numbers as continuous values; they also used the Bernoulli distribution to model the status of getting infected. 
Based on the distribution selected for modeling the data, the mean function varies accordingly, such as $\mu = \exp(\boldsymbol{w}^{T}\boldsymbol{x})$ for Poisson distribution, $\mu = \boldsymbol{w}^{T}\boldsymbol{x}$ for Gaussian distribution, and $\mu = \frac{1}{1+e^{-\boldsymbol{w}^{T}\boldsymbol{x}}}$ for Bernoulli distribution. Although GLMs have a similar regression framework to generate predictions, their specific structures and the correpsonding inference methods in different works are different, as they are elaborately designed based on various assumptions to reflect specific disease transmission processes.
% various simplified or complicated assumptions.
For example, Zhang et al. incorporated the effects of intra-regional, inter-regional, and external factors on disease risk into a unified Poisson regression-based framework to model epidemic diffusion between multiple locations \cite{zhang2015unified}. In their approach, information from prior knowledge about disease transmission is taken into consideration to specify what heterogeneous risk-related factors are involved in three aspects of disease transmission: intra-transmission, inter-transmission, and external transmission. Specifically, in the intra-transmission part of their framework, climate data (e.g., temperature, rainfall), geographical data (e.g., elevation), and demographic data (population) are combined to predict self-infections within a region; in the inter-transmission part of their framework, a diffusion matrix with a prior structure constrained by the transportation network within a region is used to describe disease transmission between locations; and in the external-transmission part of their framework, a quadratic function that has unimodal patterns is employed to model the effect of seasonal imported cases of the disease.
Similarly, Pei et al. have utilized a multivariate regression model denoted the \textbf{g}roup \textbf{s}parse \textbf{B}ayesian \textbf{l}earning model (GSBL), which is based on a transmission network, to predict disease dynamics \cite{pei2018group, pei2020active}. In contrast to \cite{zhang2015unified}, they focused on the sentinel selection problem, which arises due to limited disease surveillance resources. Sentinel selection, which is closely related to the concept of active surveillance in the domain of public health, is the selection of representative locations from all targeted locations at which to conduct disease surveillance. Pei et al. have formulated this problem as a learning process of a row-sparse disease transmission network, in which sentinels are indicated by the non-zero rows \cite{pei2018group, pei2020active}. Thus, their model uses the disease data of these sentinels and an inferred network to recover or predict the global dynamics of all target locations. In addition, the model does not use other prior knowledge about disease transmission and only uses historical case number data to infer a transmission network, so it can be more easily applied than the above Poisson regression model to various diseases and other domains.

\paragraph{Empirical Bayesian}
In contrast to GLMs, which assign a pre-defined prior distribution to the parameters of a model (i.e., a distribution that is irrelevant to the observational data), empirical Bayesian models usually estimate prior distributions from historical observations. A typical example is the semiparametric empirical Bayes framework for epidemic modeling proposed by Brooks et al. \cite{brooks2015flexible}. This framework first estimates the prior, i.e., the shape of the influenza-like illness (ILI) curve, the noise, the peak height, the peak week, and the pacing, using a set of uniform distributions over the historical observations. It then generates the underlying ILI curve of the current ILI season by linearly adjusting the piecewise quadratic curves of historical seasons using the current year's CDC baseline weekly ILI level.

\paragraph{Gaussian process}
To predict disease risks, \textbf{G}aussian \textbf{p}rocess (GP) models assume that the random variable $f(\boldsymbol{x}_i)$ in continuous domains (e.g., time or space) follow a Gaussian distribution with the mean $\mu=m(\boldsymbol{x}_i)$ and the variance $\sigma_i$, and that the joint distribution of a finite set of these variables $\boldsymbol{f}=[f(\boldsymbol{x}_1),\cdots,f(\boldsymbol{x}_M)]$ will follow the multivariate Gaussian distribution with the mean $\boldsymbol{\mu}=[m(\boldsymbol{x}_1),\cdots,m(\boldsymbol{x}_M)]$ and the covariance $\boldsymbol{\Sigma}_{ij}=\mathcal{K}(\boldsymbol{x}_i,\boldsymbol{x}_j) (i,j = 1,\cdots,M$), where the $\mathcal{K}$ is the kernel function and $M$ is the number of observations \cite{bishop2006pattern, murphy2022probabilistic}.
In GP models, given the training set $\mathbf{X}=\{{\boldsymbol{x}_1,\cdots,\boldsymbol{x}_M}\}$, we have $\boldsymbol{f}_X \sim \mathcal{N}(\boldsymbol{\mu}_X,\mathbf{K}_{X,X})$, where $\mathbf{K}_{X,X}$ is the $M \times M$ covariance matrix of the data set $\mathbf{X}$.
Given a test set $\mathbf{X}_*$, the joint distribution $p(\boldsymbol{f}_X,\boldsymbol{f}_{X_*}|\mathbf{X},\mathbf{X}_*)$ is represented as follows:
\begin{equation}
\left(\begin{array}{l}
\boldsymbol{f}_X \\
\boldsymbol{f}_{X_*}
\end{array}\right) \sim \mathcal{N}\left(\left(\begin{array}{l}
\boldsymbol{\mu}_X \\
\boldsymbol{\mu}_{X_*}
\end{array}\right),\left(\begin{array}{cc}
\mathbf{K}_{X, X} & \mathbf{K}_{X, X_*} \\
\mathbf{K}_{X, X_*}^{\top} & \mathbf{K}_{X_*, X_*}
\end{array}\right)\right).
\end{equation}
The covariance of these variables is calculated by choosing the appropriate kernel function $\mathcal{K}(\cdot,\cdot)$ and is used to describe the characteristics of processes.
Due to the inherent ability of a covariance matrix to model the similarity between data points, conventional GP models are generally used as interpolation models. However, some recent studies have extended their use by applying them to epidemic prediction tasks.
For instance, Senanayake et al. proposed a model based on GP regression that predicts influenza cases by capturing the spatiotemporal dependency of data \cite{senanayake2016predicting}. They constructed a non-linear kernel with both spatial and temporal components, and spatiotemporal covariance components, to address the challenges associated with the complicated characteristics of disease dynamics, such as temporal characteristics (i.e., periodicity, non-stationarity, and short- and long-term dependency) and spatial characteristics (i.e., the distance between locations and morphology of a region).
Zimmer and Yaesoubi proposed a GP-based framework to forecast seasonal epidemics \cite{zimmer2020influenza}. In contrast to Senanayake et al. \cite{senanayake2016predicting}, they did not design kernels to represent the dependencies between spatial locations but rather focused on exploring the temporal dependency between within-seasonal and between-seasonal time series.

\paragraph{Matrix factorization and nearest neighbor}
These methods, which are popular in the field of recommender systems \cite{koren2009matrix, sarwar2002recommender}, are also used to predict disease risks. For instance, Chakraborty et al. proposed matrix factorization with nearest-neighbor regression (MFN), which incorporates MF regression and nearest-neighbor-based regression, for ILI count prediction \cite{chakraborty2014forecasting}. In their MFN model, they integrated disease-related features, historical disease dynamics, and the disease dynamics to be predicted across time into a prediction matrix. Then, they factorized the prediction matrix as a factor-feature matrix and a factor-prediction matrix, such that the prediction matrix could be reconstructed by multiplying the factor-feature matrix and factor-prediction matrix. Subsequently, they incorporated nearest neighbor regression to correct the reconstructed prediction matrix with the $K$ nearest samples.

\paragraph{Ensemble learning}
Some studies have improved the robustness of predictions by generating them from ensemble models rather than from a single model. For instance, \cite{chakraborty2014forecasting} used an ensemble model derived from the fusion of outputs of models trained on data from various sources. Specifically, they trained multiple MFN models on data whose effects on disease dynamics had been previously studied and combined their results to give the final prediction. Similarly, \cite{zimmer2020influenza} trained various GP regression models on different features of disease-related data and aggregated these models' results to generate the final predictions.

% \cite{kandula2017subregional}
% {\color{red}(Is it necessary to mention tensor-based method in this section? Such as Yang Bo's 2017 PAMI)}

\subsubsection{Deep learning models}

\ 

\noindent
Due to the excellent ability to represent high-dimensional features in latent space and capture complex dependencies, deep learning has been widely explored and applied in the task of disease risk prediction. 
%In addition to the successful application of deep learning architectures in many different domains, such as computer vision \cite{voulodimos2018deep} and natural language processing \cite{young2018recent}, a plethora of deep learning models have been designed for disease dynamic prediction.
% {\color{teal}In \textbf{d}eep \textbf{n}eural \textbf{n}etwork (DNN) models, take the \textbf{m}ulti\textbf{l}ayer \textbf{p}erceptrons (MLPs) as an example \cite{murphy2022probabilistic}, we can generally represent the non-linear function of input variables $\boldsymbol{x}$ as $f(\boldsymbol{x};\boldsymbol{\theta})=\mathbf{W}\boldsymbol{\phi}(\boldsymbol{x}) + \boldsymbol{b}$, where $\boldsymbol{\theta}=(\mathbf{W},\boldsymbol{b})$. The $\boldsymbol{\phi}(\boldsymbol{x})$ denotes the process of feature transformation and the function is hard to be specified by hand. Therefore, it can be further expressed as another feature transformation process, i.e., $\boldsymbol{\phi}(\boldsymbol{x}) = \boldsymbol{\phi} (\boldsymbol{x};\boldsymbol{\theta}_2) = \mathbf{W}_2\boldsymbol{\phi}_2 (\boldsymbol{x}) + \boldsymbol{b}_2$, where $\boldsymbol{\theta}_2 = (\mathbf{W}_2,\boldsymbol{b}_2)$; And the feature transformation can be recursively stacked to form the deeper layers to represent more complex relationships.}
%Due to the excellent ability of deep structures to represent high-dimensional features in latent space and capture complex dependencies, 
Many sophisticated structures of \textbf{d}eep \textbf{n}eural \textbf{n}etwork (DNN) models---e.g., \textbf{c}onvolutional \textbf{n}eural \textbf{n}etworks (CNNs), \textbf{r}ecurrent \textbf{n}eural \textbf{n}etworks (RNNs), and \textbf{g}raph \textbf{n}eural \textbf{n}etworks (GNNs)---have been fully explored as a means to capture the non-linear relationships and spatiotemporal patterns of disease transmission and thereby achieve good predictive performance.
% which are more complex than MLPs
In our survey, for the simplicity of notations, we use $\hat{y}_d=f_d(\boldsymbol{x},\boldsymbol{\theta}_d)$ to represent the prediction of $y_d$ from the input feature $\boldsymbol{x}$ via a non-linear function $f_d$. Here the $f_d$ could be specified by any DNN model structure, and the $\boldsymbol{\theta}_d$ denotes the corresponding model parameters. Generally, $\boldsymbol{\theta}_d$ is optimized (over the parameter space $\boldsymbol{\Theta}_d$) using the following loss function:
% each sample with the inputs and label $(\boldsymbol{x}, y)$:
% by back-propagating the gradients
\begin{equation}
    \arg\min_{\boldsymbol{\theta}_d \in \boldsymbol{\Theta}_d} \mathcal{L}_d(\boldsymbol{\theta}_d)
\end{equation}
\begin{equation}
    \mathcal{L}_d(\boldsymbol{\theta}_d) = \ell(\hat{y}_d,y_d) = \ell(f_d(\boldsymbol{x},\boldsymbol{\theta}_d),y_d),
\end{equation}
where $\mathcal{L}_d$ is the predictive loss, quantified by the difference between the model prediction $\hat{y}_d$ and the ground truth label $y_d$. In the disease risk prediction task, various distance metrics could be used to measure the difference, such as the $\ell_1$-norm loss (mean absolute error) or $\ell_2$-norm loss (mean squared error).
% which are used to predict the disease risk.
% \textbf{m}ulti-\textbf{l}ayer \textbf{p}erception (MLP),
% As we know, MLP is the most classic and simple structure of the neural network.
% {\color{pink} In \cite{rodriguez2021deepcovid}, DeepCOVID is a real-time framework for COVID-19 forecasting containing three distinct modules that serve for a data process, prediction, and result explainability respectively. In the prediction module, they use MLP with XXX.}

%~\cite{velivckovic2017graph}
%\textcolor{CornflowerBlue}{They combine the deep neural network structure (i.e. LSTM model) and causal methods (i.e. multi-agent system) to model the non-linear relationship between the data at different spatial-temporal scales.}
%without the constraints from the compartmental models while

% add more explanations
%(Fig.~\ref{fig:IIDRNN})
\paragraph{Recurrent neural networks}
RNNs are widely used to model the temporal dependency of time series data, such as voice or text data. RNN modules are formed based on the assumption that the current output is not only related to the current input but also depends on the previous inputs. Thus, as infectious disease dynamics are a type of time series data, they can also be modeled by RNNs. For example, the \textbf{i}nteractively and \textbf{i}ntegratively connected \textbf{d}eep \textbf{r}ecurrent \textbf{n}eural \textbf{n}etwork model (I$^2$DRNN) \cite{tan2020demystifying} uses stacked RNN modules to capture spatiotemporal dependencies from heterogeneous and multiple-scale risk-related data. The model structure contains three components: (1) an input module, which is used to integrate heterogeneous (i.e., fine-, coarse-, and same-scale) data; (2) a hidden module, which is designed as a hierarchical structure to extract dependencies from different locations and the heterogeneous factors of different scales; (3) and an output module, which is used to generate final predictions based on extracted hidden features.
RNN architectures based on a gating mechanism, such as a \textbf{l}ong \textbf{s}hort-\textbf{t}erm \textbf{m}emory (LSTM) network \cite{hochreiter1997long}, have also been used in recent studies for disease risk prediction, due to their ability to preserve the long-term information of data sequences. Venna et al. proposed an LSTM-based deep learning model that consists of multiple LSTM cells sequentially stacked over time \cite{venna2018novel}. In the sequential structure, every LSTM cell takes two inputs (except the first cell, which only takes the dynamic data as input): the dynamic data at a single time point and the output of the previous cell, to generate a prediction for the next time step. That study also examined the effects of climate variables by applying the symbolic time-series approach and the effects of regions with geographical proximity by applying weighted summation to adjust the output of an LSTM to generate the final prediction.
Volkova et al. adopted two LSTM layers to learn temporal dependencies from the ILI case number and social media data, respectively, and merged their outputs via a fully connected layer to generate predicted ILI proportions \cite{volkova2017forecasting}.

\paragraph{Graph neural networks}
In contrast to RNN models, which capture the temporal dependency of sequential data, GNN models can deal with data with graphical structures \cite{zhou2020graph}. That is, a GNN model is based on a graph with a given structure and encodes structural information by passing messages between nodes of the graph. Due to this ability of GNN models to capture characteristics in graph structures, they are naturally used to capture and represent spatial patterns of disease dynamics, which can be regarded as driven by a disease transmission network.
For example, the \textbf{s}patio-\textbf{t}emporal \textbf{g}raph \textbf{n}eural \textbf{n}etwork (STGNN) \cite{kapoor2020examining} utilizes daily mobility data from Google to construct the structure of time-varying disease transmission networks. Based on their constructed network, Kapoor et al. designed two types of edges, i.e. edges between nodes within the network at the same time, and edges between nodes within the network at the current time and nodes within the network at the previous time, to characterize varying spatiotemporal dependencies driven by cross-regional human mobility and the effect of historical risk trends, respectively \cite{kapoor2020examining}.
Moreover, GNNs are not limited to modeling intuitive spatial relationships by delineating the network structure between locations; they can also be used to model the dependency between extracted features.
% with graphical structure / pair-to-pair
For instance, the \textbf{s}pectral \textbf{tem}poral \textbf{g}raph \textbf{n}eural \textbf{n}etwork (StemGNN) \cite{cao2020spectral} uses a \textbf{g}raph \textbf{c}onvolutional \textbf{n}etwork (GCN) structure to model temporal dependency to predict newly confirmed COVID-19 cases. Specifically, instead of modeling the time series in the time domain, it utilizes the \textbf{g}raph \textbf{F}ourier \textbf{t}ransform (GFT) to model inter-series correlations within the spectral domain and the \textbf{d}iscrete \textbf{F}ourier \textbf{t}ransform (DFT) to model intra-series temporal correlations within the frequency domain, and then feeds the representation of correlations into a GNN.

\paragraph{Mixed deep modules}
More recent studies have applied combinations of multiple neural network structures to model complex spatiotemporal patterns of disease transmission. These models with mixed neural network structures make full use of the aforementioned common neural network structures, i.e., RNNs, GNNs, and CNNs, to design new composite architectures serving various purposes. Usually, these architectures contain two separate modules---i.e., a spatial module and a temporal module---that are connected to form an integrated model that is subsequently optimized in an end-to-end manner to capture and model spatial and temporal dependencies simultaneously. For example, Wu et al. proposed a model which incorporates CNN, RNN, and residual structures, named CNNRNN-Res, to capture spatiotemporal dependencies in historical disease dynamics \cite{wu2018deep}. The CNN module uses an adjacency CNN filter to represent the adjacent graph of different regions, which is employed to integrate the information from neighbors. The RNN module uses a \textbf{g}ated \textbf{r}ecurrent \textbf{u}nit (GRU) to capture the temporal correlation in data. To solve the problem of overfitting, a sparse residual link is used to skip connections with some previous layers.
The \textbf{C}r\textbf{o}ss-\textbf{l}ocation \textbf{a}ttention-based GNN (ColaGNN) \cite{deng2020cola} is designed for long-term ILI prediction. It uses location-aware attention to infer the spatial influence between different regions from learned hidden features (denoting temporal dependencies), which are extracted from an RNN module. In addition, it employs a dilated convolution module to learn attributes for each node from historical disease trends and thereby capture multiple-scale local temporal dependency. Based on the above-mentioned network structure and node attributes, a graph message-passing mechanism is used to integrate the spatiotemporal information, which is then used to generate ILI predictions.
The \textbf{Hier}archical \textbf{s}patial-\textbf{t}emporal framework (HierST) \cite{zheng2021hierst} includes a temporal module that combines two time-series architectures---the long- and short-term time series network (LSTNet) and the neural basis expansion analysis for time series (N-BEATS)---to model temporal dependency; and a spatial module that contains the gated edgeGNN, which adaptively adjusts the connections of edges, and the nodeGNN, which learns the representation of node features. The novelty of this approach is also reflected by the introduction of prior knowledge of common sense to constrain the model inference. Specifically, given that the predictions for different administrative levels (i.e., country, state, and county) should be close to each other, \cite{zheng2021hierst} designed a consistency optimization objective that includes items representing the difference between predictions at different spatial scales in addition to the difference between ground truth and predictions. The \textbf{epi}demic forecasting model based on \textbf{f}unctional \textbf{n}eural \textbf{p}rocess (EPIFNP) proposed by Kamarthi et al. \cite{kamarthi2021doubt} also includes temporal and spatial modules, which are implemented by a probabilistic neural sequence encoder and a stochastic correlation graph, respectively. Instead of generating point estimates of forecast value, the EPIFNP model generates the probability distribution of prediction via a probabilistic generative process model to evaluate the uncertainty of prediction.
The population-level disease prediction model (named PopNet) proposed by Gao et al. assumes that an undirected disease transmission network drives disease dynamics \cite{gao2022popnet}. PopNet learns the connection structure of this network by using population and geographical distance to calculate the similarity of each pair of locations. Then, based on the learned network structure, PopNet uses two graph attention networks (GATs) to obtain node embedding from real-time disease data and updated disease data (i.e., revised data released after the initial release), respectively. Subsequently, PopNet fuses these two kinds of node embeddings by \textbf{s}patial \textbf{l}atency-aware \textbf{att}ention (S-LAtt) and \textbf{t}emporal \textbf{l}atency-aware \textbf{att}ention (T-LAtt), sequentially. S-LAtt uses a feature similarity-based attention mechanism and considers the marginal effects of time latency on final predictions to learn the edge weights between pairs of nodes together to update node embedding. T-LAtt uses GRU networks to learn temporal dependency. Finally, PopNet concatenates the learned node embeddings to generate final predictions.

% Besides the combination of various deep learning modules, several general frameworks are taken for more specific purposes in the task of epidemic prediction.
%And it is also applied to model transformation between multimodal data.
%which exceeds the prediction on one time step.
%long-term prediction.
\paragraph{Encoder--decoder}
The encoder--decoder framework is a general framework that consists of different deep modules to manage sequential data and was initially used in machine translation \cite{cho2014learning}. As trends in disease dynamics exhibit a sequential dependency that is similar to that of sentences of text, an encoder--decoder framework with an RNN structure can also be used to predict epidemic trends within a given time period \cite{adhikari2019epideep, cui2021into}. 
Adhikari et al. proposed EpiDeep to predict weighted ILI (wILI) using an encoder--decoder framework, together with deep clustering components \cite{adhikari2019epideep}. EpiDeep uses an LSTM-based encoder to encode an input influenza sequence as latent variables that contain temporal information, and a deep clustering component (an \textbf{i}mproved \textbf{d}eep \textbf{e}mbedded \textbf{c}lustering (IDEC) module \cite{guo2017improved}) to learn the embedding of the existing observed epidemic trend in the current season whose trend is to be predicted, and then clusters this embedding with the most similar epidemic trends in historical seasons. EpiDeep also uses this approach to learn and cluster the embedding of full-length historical trends. Next, it learns a mapping function to map the embedding of the incomplete sequence to the space of the full-length sequence. Finally, EpiDeep uses a decoder to predict the future sequence of the epidemic trend in the current season by taking the mapped clustering embedding and the encoded trend (both are in the current season) as inputs.
Cui et al. also used an encoder--decoder framework to predict the dynamics of COVID-19 pandemic \cite{cui2021into}. The encoder component employs several CNN modules with different kernel sizes to extract temporal features within multiple time ranges from the data of case numbers and regional visitor counts. They designed a graph-based module that characterizes spatial patterns by modeling human mobility and infection processes in each range in which graph structure is learned from data by the attention mechanism, and then fuses features learned from each range into one feature via a multi-headed self-attention mechanism. The decoder component employs a temporal embedding module to embed the case numbers and death numbers, and the obtained embedding is fed into the multi-head attention layers with the output of the encoder. Finally, the features obtained from the encoder and decoder are passed through a multilayer perceptron (MLP) to generate a final prediction.

% utilizing the patterns learned from other models or data to predict.
\paragraph{Transfer learning}
However, knowledge on the properties and transmission of emerging diseases may be ambiguous, and there may be a paucity of observational data on epidemic trends. Additionally, the data for regions with inadequate systems for surveillance and reporting of infectious diseases may be incomplete and noisy. Therefore, it is of interest to determine how to use abundant and high-quality data on similar diseases to the disease of interest or for regions with similar characteristics to the region of interest to facilitate the prediction of a given disease epidemic. One approach used is \textbf{t}ransfer \textbf{l}earning (TL) architecture, which consists of one source task and one target task, and is designed to transfer knowledge learned from the previous model to enable the learning of the target task \cite{pan2009survey}.
Similarly, the COVID augmented ILI deep network (CALI-NET) is a \textbf{h}eterogeneous \textbf{t}ransfer \textbf{l}earning (HTL) framework for COVID-ILI forecasting  \cite{rodriguez2021steering} that applies the EpiDeep \cite{adhikari2019epideep} model as the source model to learn representations of temporal dependency from historical wILI data. \cite{rodriguez2021steering} designed the \textbf{C}OVID-\textbf{a}ugmented \textbf{e}xogenous \textbf{m}odel (CAEM) to encode representations of spatiotemporal features of exogenous data signals of COVID-19 by Laplacian regularization of a geographical adjacent matrix and a GRU module, and used these encoded representations in the target model. They also designed a \textbf{k}nowledge \textbf{d}istillation (KD) loss, which consists of the hint loss between the mapped representations from source and target models, and the imitation loss between source predictions and ground truth to ensure effective knowledge transfer.

% By mapping the representations obtained from the source model and CAEM into the same latent space, they
% apply the patterns captured from the rich wILI data of historical epidemic seasons to facilitate the prediction of COVID-19

\vspace{0.2cm}

The above-described studies demonstrate that previously developed deep learning-based models use spatial and temporal modules, which are designed to capture disease patterns over space and time, respectively.
The relationship between spatial information is typically modeled by a GNN module. However, using a GNN module is challenging, due to its unknown network structure. Some approaches pre-build connections in a network by using proxy data, such as human mobility data \cite{kapoor2020examining}, population and geographical distance data \cite{gao2022popnet}, and regional demarcation data \cite{rodriguez2021steering}, and then feed the network into a GNN module or constrain model learning by applying graph Laplacian regularization. If there is a lack of related data, a graph structure can be constructed by using the attention mechanism to infer the weights of edges in a network structure during model optimization \cite{deng2020cola, zheng2021hierst, gao2022popnet, cui2021into}.
Temporal patterns are modeled by mining the temporal dependency of time series data using various RNN structures, such as a classic RNN \cite{deng2020cola}, a GRU \cite{wu2018deep, rodriguez2021steering, kamarthi2021doubt, gao2022popnet}, LSTM \cite{adhikari2019epideep}, or an LSTNet/N-BEATS \cite{zheng2021hierst}.
However, in addition to using an RNN and GNN to characterize temporal and spatial patterns, respectively, these neural network modules can be flexibly designed to serve these purposes. For instance, in addition to RNN modules being used to extract temporal dependencies, CNN modules (such as temporal convolution modules \cite{cui2021into} or dilated CNNs \cite{deng2020cola, gao2022popnet}) and GNN modules (such as spectral temporal GNNs \cite{cao2020spectral}) can also be used for this purpose; and in addition to GNN modules being used to capture spatial dependencies, a modified CNN module can also be used for this purpose \cite{wu2018deep}.

% only capturing specific temporal spatial patterns. Actually, the convolution modules can also be used to extract the temporal patterns for more advanced purposes,
% such as temporal convolutional Networks (TCN) which XXX, the convolution over time dimension \cite{cui2021into};
% dilated convolution which is used to extract temporal patterns at multiple time scales \cite{gao2022popnet}.

%-----------------------------------
%Besides constructing the network structure with mobility data, the attention mechanism is also used to infer the network structure, especially in the case of the absence of data directly describing the disease transmission patterns.
%The framework of the Cola-GNN model can be found in Fig.~\ref{fig:ColaGNN}.
% by constructing a graph neural network in which the nodes represent the locations, the relationship between nodes is assigned by the learned attention matrix and initial features of nodes are denoted by the learned hidden representation of temporal dependency.

\subsubsection{Discussions: advantages and limitations}

\ 

\noindent
The above-mentioned studies show that many data-driven machine learning methods have been developed for infectious disease risk prediction. Their excellent predictive performance may be due to the following aspects.
\begin{itemize}
    \item The sophisticated modules designed for spatial and temporal characterization can extract complex hidden representations and learn complex non-linear relationships from abundant data sources to capture spatiotemporal disease transmission patterns.
    \item The above-described data-driven models are based on supervised learning algorithms that enable the inference of model parameters that fit well with the data.
    % They have various and flexible model structures which enable better capacity to fit the training data based on the XXX.
\end{itemize}

However, most of the above-described advanced machine-learning methods are not designed for the modeling of diseases with complex transmission environments and conditions, such as malaria. Thus, these methods may not perform well when they are applied to model such diseases because they do not fully consider and utilize disease-related knowledge on a given disease. Furthermore, it is difficult to quantify the effects of various risk-related factors on predicted transmission intensity or risk in an interpretable way, so modeling results may not provide clear guidance for decision-makers on implementing time/location/factor-specific measures in response to potential risks or outbreaks.
% which are important to accurately predict the future trends of disease dynamics
% {\color{red}using either the mechanism-based models or data-driven models along (Please see if this change expressed what you want to say)}

\subsection{Epidemiology-inspired machine learning}\label{Epidemiology-inspired machine learning}
Although data-driven machine learning approaches greatly improve the accuracy of disease risk prediction, they still struggle to provide insights to facilitate disease control. To overcome this drawback, epidemiological models have been re-visited and integrated with machine learning methods. Epidemiological models mathematically depict a disease transmission process based on domain experts' understanding of the disease's biological characteristics. Thus, each parameter and the overall structures of epidemiological models have a clear epidemiological meaning. However, the structures of epidemiological models are typically based on relatively simplified assumptions, so these models may struggle to provide sufficiently accurate predictions. Conversely, data-driven machine learning models can fit training data very well and generate accurate predictions, but in some cases the physical meaning of learned patterns is ambiguous and thus cannot effectively support public health decision-making. Therefore, a key question in disease risk prediction modeling is how to exploit the complementary strengths of data-driven models and epidemiological models to obtain modest explanatory power while utilizing their strong representation ability to determine complex dependencies. Driven by this question, a large body of literature has investigated the potential of combinations of epidemiological models and data-driven machine-learning models. In this paper, we denote this type of model "epidemiology-inspired machine learning" and we divide it into two classes: (1) epidemiological parameterization and (2) epidemiology-embedded learning. In the following, we first provide some preliminary information on epidemiological models. Then, we introduce previous studies by categorizing their epidemiology-inspired machine learning models and describing how they combine epidemiological prior knowledge with machine learning methods. A summary of the epidemiology-inspired machine learning models is given in Table \ref{tab:Epidemiology-inspired machine learning}.

\begin{table}[htbp]
  \centering
  \caption{Summary of epidemiology-inspired machine learning models.}
  \label{tab:Epidemiology-inspired machine learning}%
  \scalebox{0.75}{
    \begin{tabular}
    {m{3.8cm}<{\centering}|m{3.8cm}<{\centering}|ccm{3cm}<{\centering}m{4cm}<{\centering}}
    \toprule
    \textbf{Categories} & \textbf{How to combine} & \textbf{Ref.} & \textbf{Targeted diseases} & \textbf{Epidemiological components} & \textbf{Data-driven components} \\
    \midrule
    \multirow{10}[20]{2cm}{\centering Epidemiological parameterization} & \multirow{8}[16]{2cm}{\centering Inferring epidemiological parameters from data} & \cite{shaman2012forecasting, shaman2013real} & Influenza & Humidity-driven SIRS model & EAKF/PF (Data assimilation) \\
    \cmidrule{3-6}          &       & \cite{yang2015inference} & Infuenza & Four types of compartmental models & EAKF/PF (Data assimilation) \\
    \cmidrule{3-6}          &       & \cite{pei2018forecasting} & Infuenza & Metapopulation compartmental model & EAKF/PF (Data assimilation) \\
    \cmidrule{3-6}          &       & \cite{tizzoni2012real} & A/H1N1 & GLEaM & Monte Carlo maximum likelihood analysis \\
    \cmidrule{3-6}          &       & \cite{zhang2017forecasting} & Influenza & GLEaM & Monte Carlo maximum likelihood analysis \\
    \cmidrule{3-6}          &       & \cite{zou2020epidemic} & COVID-19 & SuEIR model & Loss function with logarithmic-type MSE \\
    \cmidrule{3-6}          &       & \cite{wang2021bridging} & COVID-19 & Spatiotemporal-SuEIR & AutoODE \\
    \cmidrule{3-6}          &       & \cite{wang2018inferring} & Airborne disease & Metapopulation SIR model & Non-negative network inference model with power-law distribution and data priori \\
    \cmidrule{2-6}          & \multirow{2}[4]{2cm}{\centering Modeling epidemiological parameters} & \cite{arik2020interpretable} & COVID-19 & Improved SEIR model & Generalized additive model \\
    \cmidrule{3-6}          &       & \cite{baek2020limits} & COVID-19 & Stochastic SIR process & Mixed effects model \\
    \midrule
    
    \multirow{8}[16]{2cm}{\centering Epidemiology-embedded learning} & \multirow{2}[4]{2cm}{ \centering Epidemiological guides} & \cite{shi2020inference} & Malaria & EIR/VCAP & Nonlinear stochastic model \\
    \cmidrule{3-6}          &       & \cite{liu2023assessing} & Malaria & NGM   & Multiplevariate regression with non-linear parameters \\
    \cmidrule{2-6}          & \multirow{6}[12]{2cm}{\centering Epidemiological regularization and constraints} & \cite{hua2018social} & COVID-19 & SEIR model & Social media based simulation model \\
    \cmidrule{3-6}          &       & \cite{kargas2021stelar} & COVID-19 & SIR model & Spatiotemporal tensor factorization \\
    \cmidrule{3-6}          &       & \cite{osthus2019dynamic} & Influenza & SIR model & Dynamic Bayesian \\
    \cmidrule{3-6}          &       & \cite{gao2021stan} & COVID-19 & SIR model & GAT and GRU \\
    \cmidrule{3-6}          &       & \cite{wang2022causalgnn} & COVID-19 & SIRD model & Dynamic attention-based GCN \\
    \cmidrule{3-6}          &       & \cite{wang2019defsi, wang2020tdefsi} & Influenza & SEIR model & LSTM \\
    
    \bottomrule
    \end{tabular}
    }
\end{table}%

\subsubsection{Introduction to epidemiological models}\label{Introduction to epidemiological models}

\ 

\noindent
In the 20th century, many epidemiological models were developed to mathematically depict the process of infectious disease transmission based on the understanding and knowledge of disease characteristics and transmission modes. These models are also known as mechanism-based models, compartmental models, or physics-based models. In these models, a studied population is usually divided into several compartments representing different disease statuses, and a set of rules is designed to describe the transition between these statuses. These models can be subdivided based on the granularity level of modeling (from coarse to fine) into three types \cite{liu2020computational, adiga2020mathematical}: (1) compartmental models at the population level; (2) compartmental models at the meta-population level; and (3) agent-based at the individual level.
% \paragraph{Population level}
Compartmental models at the population level usually include a set of differential equations (DEs) that depict the dynamics of state variables and thereby quantitatively represent disease risk. These models comprise a representative and classical group of epidemic models that are used to mathematically depict disease transmission. Various DEs have been developed for a wide range of infectious diseases, such as influenza, malaria, dengue, Aids, and COVID-19. These models assume that a disease transmission environment is homogeneous, i.e., individuals are mixed evenly within the environment and those with the same status have the same probability of moving  from their current status to another status. 
In the following section, we introduce various compartmental models at the population level that have been constructed for diseases that propagate in different ways, e.g., respiratory diseases and vector-borne diseases.
% \paragraph{Respiratory diseases}

The \textbf{s}usceptible--\textbf{i}nfected--\textbf{r}ecovered (SIR) model is a classic compartmental model \cite{kermack1927contribution}. It has a simple structure with three statuses (susceptible, infected, and recovered) and two parameters (effective contact rate $\beta$ and recovery rate $\gamma$), and is widely used to model the dynamics of infectious diseases \cite{tolles2020modeling}, especially respiratory diseases, such as influenza, ILI, and COVID-19.
% Although the SIR model can characterize some epidemic trends and it is easy to simulate and analyze due to the simplicity of its formulation, it is still hard to handle more complex disease processes, such as the latent period of humans before an individual becomes contagious after they contact the infectious sources.
Following the development of the SIR model, many other compartmental models with more sophisticated structures were designed to describe more complex scenarios, such as \textbf{s}usceptible--\textbf{e}xposed--\textbf{i}nfected--\textbf{r}ecovered (SEIR) models, which consider the latent period of a disease \cite{hethcote1991some, li1995global} and use the parameter $\beta$ to represent the probability of an individual entering the incubation period after being in contact with an infectious individual, and the parameter $\alpha$ to represent the probability of an individual leaving the latent period; and the \textbf{s}usceptible--\textbf{e}xposed--\textbf{i}nfected--\textbf{r}ecovered--\textbf{d}eath (SEIRD) model \cite{maugeri2020estimation, piccolomini2020preliminary}, which considers deaths due to disease.
% \paragraph{Vector-borne diseases}
Many variations of compartmental models have been developed for vector-borne diseases, such as malaria and dengue, to depict disease transmission between vectors (e.g., mosquitoes) and humans \cite{mandal2011mathematical, andraud2012dynamic}.
For instance, the Ross model \cite{ross1908report} is the most fundamental model to describe vector-borne diseases, while the Macdonald model \cite{macdonald1957epidemiology} is based on the Ross model but also considers the latent status of vectors. In addition to considering the latent status of vectors, the Anderson and May model \cite{anderson1992infectious} considers the latent status of humans, and \textbf{s}usceptible--\textbf{l}atent--\textbf{i}nfected--\textbf{r}ecovered (SLIR) models \cite{yang2000assessing, yang2000malaria} consider the recovered status of humans in terms of acquired immunity.

% \subsubsection{Meta-population level}
However, sometimes the homogeneous-mixing assumption does not accurately reflect the real situation of disease transmission because individuals in a host group may have different characteristics, such as different susceptibilities to disease and abilities to recover from infection. These characteristics significantly influence disease spread throughout a population and also determine how epidemic interventions should be enacted. Therefore, in addition to models based on the assumption that disease spreads occurs in a homogeneous environment (i.e., that individuals have the same probability of coming into contact with each other and of moving from one status to another), many models---i.e., compartmental models at the meta-population level---have been developed that are not based on this assumption; instead, they are based (to some extent) on a heterogeneous assumption. Studies have divided populations into subgroups and designed model structures according to different population properties, such as age structures \cite{castillo1989epidemiological, filipe2007determination}, geographical distributions \cite{cosner2009effects, prosper2012assessing, pei2020initial}, and human behavioral patterns \cite{jacquez1988modeling}.

% \subsubsection{Individual level}
As mentioned, mechanistic models at the meta-population level consider the heterogeneity of subgroups of a whole population. However, their characterization of the heterogeneity of population traits is still limited because of the low resolution of subgroup partitions. Contact between hosts and hosts, or between hosts and vectors, is the natural way in which infectious diseases are transmitted in the real world. Thus, agent-based models are usually built on a network constructed at the individual level and simulate interactions between individuals, such that they model disease transmission in more realistically than mechanistic models at the meta-population level.
EpiSims, proposed by Eubank et al., is an agent-based simulation tool for modeling disease spread caused by human mobility \cite{eubank2004modelling}. EpiSims simulates the physical contact patterns of humans by constructing a bipartite social contact network that consists of two types of vertices: individual vertices and location vertices. Compared with the results of compartmental models at the population and meta-population levels, the simulations generated by agent-based models are closer to the real-world situation because the characteristics of these models' networks are similar to those of real networks.
% For example, they found that constructed network between people has the small-world feature while the location network is scale-free and these network characteristics could drive rapid disease transmission among people over space.
Similarly, Hoertel et al. developed a stochastic agent-based microsimulation model for modeling the COVID-19 epidemic in France \cite{hoertel2020stochastic}.
% This agent-based model includes three components: (1) synthetic populations generated with realistic demographic characteristics; (2) a social contact network that describes the connections between individuals; and (3) a disease model which depicts the disease transmission through edges.
The two above-described studies show that fine-grain agent-based models enable the flexible setting of interventions and can help to reveal potentially effective intervention strategies.

\subsubsection{Epidemiological parameterization}
% \subsubsection{Data-informed mechanism models}
%fitted model pertinent to the data

\ 

\noindent
Epidemiological parameterization uses or improves existing mechanism-based models (e.g., SIR and SEIR models) as the basis for predicting disease dynamics. The initial values and epidemiological parameters of compartmental models that are described by ordinary DEs (ODEs) are crucial for determining disease dynamic. In contrast to traditional mechanism-based models, which use given or fixed epidemiological parameters, epidemiological parameterization models use disease-related data to estimate model variables and epidemiological parameters in compartmental models.
% {\color{cyan}In existing work, many }
% This type of model can be further classified into two classes: one is inferring the model parameter based on observed data with the optimization process of machine learning; another is directly calculating the parameters by collected data with prior knowledge.
% For the former type of model,
Many machine learning approaches are used to infer model variables and epidemiological parameters. The relatively simplified structures (compared with real systems) and inaccurate parameterization of initial values and parameters in mechanism-based or compartmental models may cause them to generate biased predictions (i.e., predictions that do not reflect the real situation), so inference approaches try to compensate for biased predictions to some extent by balancing model predictions with observed data.

\paragraph{Inferring epidemiological parameters from data}
Data assimilation techniques, which are widely applied in atmospheric and oceanic sciences and in numerical weather forecasting \cite{wang2000data}, aim to utilize observations to optimize mechanism-based models. Thus, they have also been applied in disease dynamic prediction \cite{shaman2012forecasting, shaman2013real, yang2015inference, pei2018forecasting}.
In a set of data assimilation models used for epidemic prediction, the \textbf{K}alman \textbf{f}ilter (KF) and its variants \cite{welch1995introduction, anderson2001ensemble} and \textbf{p}article \textbf{f}ilter (PF) \cite{arulampalam2002tutorial} methods have been used to estimate model statuses.
For instance, Shaman and Karspeck applied data-assimilation techniques to the problem of influenza forecasting and generated retrospective ensemble forecasts of influenza seasons from 2003 to 2008 in New York City, USA \cite{shaman2012forecasting}. They proposed the SIRS--EAKF framework, which uses the \textbf{e}nsemble \textbf{a}djustment \textbf{K}alman \textbf{f}ilter (EAKF) \cite{anderson2001ensemble}) and a PF \cite{arulampalam2002tutorial} to assimilate the observations of infections (i.e., estimates of influenza infections from Google Flu Trends) into the \textbf{s}usceptible--\textbf{i}nfectious--\textbf{r}ecovered--\textbf{s}usceptible (SIRS) model \cite{shaman2010absolute}), which is a humidity-forced compartmental model.
The SIRS--EAKF framework can estimate the posterior of probabilistic distributions of system states (i.e., susceptible populations $S_t$ and infected populations $I_t$) and epidemiological parameters (e.g., the mean infectious period $D$, the average duration of immunity $L$, and the maximum and minimum of daily basic reproductive number $R_{0max}$ and $R_{0max}$) in the used SIRS model.
In \cite{shaman2010absolute},  Shama et al. represented model states and epidemiological parameters by a set of variables $Z_t=(S_t, I_t, R_{0max}, R_{0min}, L, D)$. Then the posterior of $Z_t$ can be represented as follows:
\begin{equation}
p(Z_t|y_t,y_{t-1},\cdots) \propto p(y_t|Z_t)p(Z_t|y_{t-1},\cdots),
\end{equation}
where the first term on the right-hand side is the likelihood of observational disease risk given states and parameters, while the second term is the prior distribution of the states and parameters. For KF, these two terms are assumed to be Gaussian distributions; in contrast, for PF, these two terms are not under these assumptions.
Subsequently, Shaman et al. used similar data-assimilation techniques to generate weekly influenza forecasts for the influenza season in 2012 and 2013 in 108 cities in the USA \cite{shaman2013real}.
Yang et al. tested four types of compartmental models and two types of filter models in their model-data assimilation framework and analyzed the epidemiological characteristics of influenza dynamics from the 2003--2004 season to the 2012--2013 season in 115 cities in the USA \cite{yang2015inference}.
Pei et al. developed a model-data assimilation framework based on a metapopulation compartmental model to accurately predict the spatial spread of influenza \cite{pei2018forecasting}. In this metapopulation compartmental model, which is based on a humidity-driven SIRS model \cite{shaman2010absolute} that they had  used in their previous studies \cite{shaman2012forecasting, shaman2013real, yang2015inference}, they divided a population into different groups in terms of geographical locations (i.e., different states), and incorporated two types of human mobility (i.e., fixed commuting flows and irregular movement of visitors).

% {\color{cyan} Some simulation-based models which are given the epidemiological parameters or mining the parameters with limited data points are also used to describe and predict infectious disease transmission.}
Balcan et al. proposed the \textbf{gl}obal \textbf{e}pidemic \textbf{a}nd \textbf{m}obility (GLEaM) computational model for simulating infectious disease transmission \cite{balcan2009multiscale, balcan2010modeling}. The GLEaM model is a global model based on a stochastic compartmental model at the meta-population level (i.e., a SLIR model considering both symptomatic and asymptomatic infections) and incorporates multiscale human mobility (short-range commuting and long-range airline flows) to effectively capture disease transmission patterns.
Tizzoni et al. used the GLEaM model to model the disease transmission of 2009 H1N1 influenza and utilized the Monte Carlo maximum likelihood method to estimate some parameters \cite{tizzoni2012real}.
Similarly, Zhang et al. proposed an epidemic computational framework based on the GLEaM model \cite{zhang2017forecasting}. The computation is performed via three steps: (1) microblogging data from Twitter and surveillance data are used to estimate initial infections; (2) epidemiological parameters are searched in four-dimensional space by running Monte Carlo simulations with selected sampling points, and the GLEaM model is used to generate simulations; (3) a set of best-fit models is selected by using a multi-model information approach, which minimizes the loss of information (which is calculated by the \textbf{A}kaike \textbf{i}nformation \textbf{c}riterion (AIC)).

Aside from data-assimilation methods and simulation-based methods, some machine learning approaches are proposed to estimate the model states and epidemiological parameters. In these works, the loss function is generally formulated as the difference between states simulated using epidemiological models and the ground truth of these states. The general formulation of such loss function can be represented as follows:
\begin{equation}
    \mathcal{L}_e(\boldsymbol{\theta}_e) = \ell(\hat{\boldsymbol{y}}_e,\boldsymbol{y}_e) = 
    \ell(f_e(\boldsymbol{\theta}_e, \boldsymbol{s}_0), \boldsymbol{y}_e),
\end{equation}
where $\boldsymbol{\theta}_e$ (in the parameter space $\boldsymbol{\Theta}_e$) denotes epidemiological parameters (e.g., contact rate and recovery rate) in a given epidemiological model, $\boldsymbol{y}_e$ denotes the ground truth of the target variable (usually are model states with records, e.g., infected case number and death number), $\hat{\boldsymbol{y}}_e$ denotes predictions on the target variable, $\boldsymbol{s}_0$ denotes the initial value of model states, and $f_e$ denote the given epidemiological model, which is generally described by an ODE. With such an ODE representation, $\hat{\boldsymbol{y}}_e$ can be calculated using model parameters and initial values.
% $\ell$ denotes the way to calculate the error between predicted values and ground truth.
Given the above loss function, the optimal model parameters can be inferred by minimizing the loss:
% \begin{equation}
% \hat{\boldsymbol{\theta}}_e = \arg\min_{\boldsymbol{\theta}_e \in \boldsymbol{\Theta}_e } \mathcal{L}_e(\boldsymbol{\theta}_e; \mathbf{s}).
% \end{equation}
\begin{equation}
\hat{\boldsymbol{\theta}}_e = \arg\min_{\boldsymbol{\theta}_e \in \boldsymbol{\Theta}_e } \mathcal{L}_e(\boldsymbol{\theta}_e).
\end{equation}
For example, Zou et al. formulated a loss function with a logarithmic-type \textbf{m}ean \textbf{s}quare \textbf{e}rror (MSE)  \cite{zou2020epidemic}. Based on this loss function, parameters can be optimized by the general gradient-based optimizer.
Moreover, Zou et al. also developed a novel compartmental model, named the SuEIR model---an improved \textbf{SEIR} model that considers a scenario of \textbf{u}ntested or \textbf{u}nreported cases of COVID-19---and trained it with their machine learning approach.
Wang et al. formulated a similar loss based on MSE to estimate the parameters of epidemiological models. Based on this loss function, they formalized the learning procedure as the AutoODE algorithm, which infers the model parameters of mechanism-based models by an automatic differentiation method.
In addition, based on a case study on the forecasting of COVID-19 dynamics, they proposed the spatiotemporal SuEIR model, which is an extension of the SuEIR model \cite{zou2020epidemic} that better models spatiotemporal patterns of COVID-19 spread.
The power-law \textbf{d}egree and \textbf{d}ata \textbf{p}riori jointly \textbf{r}egularized non-negative network \textbf{i}nference ($D^{2}PRI$) approach of Wang et al. \cite{wang2018inferring} is based on a SIR model at the meta-population level. As this model regards the infectious interactions between individuals at different locations as a transmission process in a disease propagation network, Wang et al. \cite{wang2018inferring} formulated the parameter inference of edge weights in the network and disease transmission rate in the SIR model as an integrated network inference problem. Moreover, according to prior knowledge of network structure, i.e., the power-law distribution of node degree and the features extracted from mobility-related data, they designed corresponding regularization items to constrain the parameter inference.

\paragraph{Modeling epidemiological
parameters}
In contrast to models that infer values or probabilistic distributions of model parameters from observations, other models estimate the variation of epidemiological parameters and formulate them as functions of covariates. For instance, Arik et al. proposed the use of time-varying functions to model parameters \cite{arik2020interpretable}. That is, they used an improved compartmental model that is based on the SEIR model: instead of using the static epidemiological parameters in the traditional compartmental model, they used learnable functions to estimate parameter values from various covariates, which enable parameter values to vary over time. Specifically, they used the generalized additive model to encode the effects of covariates on epidemiological parameters.
Baek et al. predicted the disease dynamics of multiple regions by using a stochastic SIR model \cite{baek2020limits}. This stochastic model employs a mixed-effects model that incorporates a random-effects term within each region and a fixed-effects term between different regions to encode the effects of static and time-varying covariates on the disease transmission rate.

% For the latter type of model, instead of modeling the disease dynamics and inferring the epidemiological parameters by machine learning approaches simultaneously, they usually separate this process as a two-stage approach which the first step is estimating relevant parameters from known risk-related factors and the second step is modeling disease dynamics by directly using the calculated parameters.

\subsubsection{Epidemiology-embedded learning}
% \subsubsection{Mechanism-guided machine learning models}

\ 

\noindent
In contrast to epidemiological parameterization, epidemiology-embedded learning focuses on using machine learning models to predict disease dynamics directly, while using mechanism-based models to guide, regularize, or constrain the machine learning models.

\paragraph{Epidemiological guides}
Some studies have utilized the epidemiological concept to guide the construction of model structures. For instance, rather than directly describing and predicting malaria dynamics by using ODEs, Shi et al. constructed a nonlinear stochastic model based on the formula for vectorial capacity (VCAP) and entomological inoculation rate (EIR), which are epidemiological concepts derived from corresponding ODEs that describe malaria transmission \cite{shi2020inference}. VCAP is defined as the daily rate of future inoculations from mosquitoes to humans caused by a currently infected human case \cite{smith2004statics}, whereas EIR is defined as the number of infectious bites received from mosquitoes per day by a human \cite{smith2004statics}. Based on the epidemiological meaning of local transmission, the local infections at time $t$ can be formulated as a function involving the VCAP/EIR and the infections at time $t-1$. Then, by considering the effects of cross-regional transmission, Shi et al. used periodic function modeling to depict the periodic transmission patterns \cite{shi2020inference}. Thus, their nonlinear stochastic model consists of the items of local infections and imported infections.
Liu et al. developed a multivariate regression model based on a next-generation matrix of a meta-population vector--human compartmental model to predict the malaria risk in multiple locations\cite{liu2023assessing}. The next-generation matrix \cite{diekmann1990definition, van2002reproduction}, consisting of the non-linear relationships between epidemiological parameters that can be derived from a compartmental model, represents the change in model variables from one time step to the next time step, thereby enabling the prediction of disease dynamics.
% add other works of Benyun.

\paragraph{Epidemiological regularization and constraints}
Some studies have added epidemiological constraints and regularizations, which are derived from compartmental models, to standard objective functions of supervised machine learning models to aid model parameter optimization.
Hua et al. proposed the \textbf{s}ocial \textbf{m}edia based \textbf{s}imulation (SMS) model for influenza dynamics prediction \cite{hua2018social}. This model incorporates two learning spaces: the social media space, which is designed to identify individuals' health statuses from social media posts; and the epidemiological simulation space, in which a transmission network is built to simulate disease propagation between individuals. These two spaces are linked by minimizing the loss in terms of the inconsistency between the health status at the population level, which is obtained from the social media space and the simulation space.
Kargas et al. applied epidemiological constraints in tensor factorization approaches to predict disease dynamics \cite{kargas2021stelar} by devising \textbf{s}patio-\textbf{t}emporal tensor factorization with \textbf{e}pidemio\textbf{l}ogic\textbf{a}l \textbf{r}egularization (STELAR). A tensor is an intuitive and natural structure used to represent and preserve the complex structure of high-dimensional data, especially spatiotemporal data with multiple risk-related factors. Tensor factorization is usually employed for dimensionality reduction and data decomposition rather than to predict disease transmission dynamics. STELAR enables the prediction of long-term epidemic trends by the addition of the latent epidemiological regularization of the SIR model into a standard tensor factorization method, i.e. \textbf{c}anonical \textbf{p}olyadic \textbf{d}ecomposition (CPD).
% {\color{red}
% The loss with epidemiological regularization and constraint is shown in Eq.~(\ref{STELAR}):
% \begin{equation}\label{STELAR}
% \begin{aligned}
% \min _{\substack{\boldsymbol{\beta}, \mathbf{B}, \mathbf{C}, \mathbf{i}, \mathbf{i}}} & \|\underline{\mathbf{X}}-\llbracket \mathbf{A}, \mathbf{B}, \mathbf{C} \rrbracket\|_F^2+\mu\left(\|\mathbf{A}\|_F^2+\|\mathbf{B}\|_F^2+\|\mathbf{C}\|_F^2\right) + \nu \sum_{k=1}^K \sum_{t=1}^L\left(c_{t, k}-\beta_k S_k(t-1) I_k(t-1)\right)^2 \\
% \text { s. t. } & \mathbf{A} \geq \mathbf{0}, \mathbf{B} \geq \mathbf{0}, \mathbf{C} \geq \mathbf{0}, \boldsymbol{\beta} \geq \mathbf{0}, \boldsymbol{\gamma} \geq \mathbf{0}, \mathbf{s} \geq \mathbf{0}, \mathbf{i} \geq \mathbf{0}, \\
% & S_k(t)=S_k(t-1)-\beta_k S_k(t-1) I_k(t-1) \\
% & I_k(t)=I_k(t-1)+\beta_k S_k(t-1) I_k(t-1) -\gamma_k I_k(t-1), \\
% & s_k=S_k(0), i_k=I_k(0).
% \end{aligned}
% \end{equation}
% In the above equation, the first and second items of loss are the standard loss of CPD, and the third item denotes the errors between the predicted new infections at time $t$ and the .
% }
Osthus et al. proposed a \textbf{d}ynamic \textbf{B}ayesian (DB) influenza forecasting approach that models discrepancies between mechanistic model-generated simulations and observations \cite{osthus2019dynamic}. This approach assumes that the uncertainty of prediction cannot be fully explained by observational noise and therefore models a wILI as the sum of three items: the logit of the infections that are described by the SIR model, a common discrepancy item for all influenza seasons, and a specific discrepancy item for each influenza season.

%The objective function is provided in the Eq.~\ref{Eq:STELAR} (\cite{kargas2021stelar}). The first and second items are the classic optimization goal of the nonnegative CPD problem (data fitting term and regularization term), and the last item is the constraint from the compartmental model.

%Some work is based on the graph attention network with and dynamics-based loss term to capture the spatial-temporal features from both real-world claims data and disease case counts data.

% deep learning model
Some deep learning models, such as GNNs \cite{gao2021stan, wang2022causalgnn} and RNNs \cite{wang2019defsi, wang2020tdefsi}, also incorporate mechanistic models to constrain the learning of model structures and parameters, such that they effectively fit the realistic situation of disease transmission.
In addition to the prediction loss, some of the aforementioned methods further introduce the epidemiological loss to ensure that the prediction of deep learning models is consistent with the dynamics described by the epidemiological model. The general formulation of the loss function is shown as follows:
\begin{equation}
    \mathcal{L} = \mathcal{L}_d (\boldsymbol{\theta}_d) +\mathcal{L}_e (\boldsymbol{\theta}_e) = \ell(\hat{y}_d,y_d) + \ell(\hat{\boldsymbol{y}}_e,\boldsymbol{y}_e),
\end{equation}
where $\mathcal{L}_d$ denotes the prediction loss to ensure the prediction accuracy of the deep learning model, and $\mathcal{L}_e$ denotes the epidemiological-constrained loss. The approaches to introduce $\mathcal{L}_e$ vary in different works. For example, \textbf{s}patio-\textbf{t}emporal \textbf{a}ttention \textbf{n}etwork (STAN) is a GAT model with epidemiological constraints that is designed for long-term pandemic prediction \cite{gao2021stan}. In the network, nodes denote different locations and have both static features (latitude, longitude, population size, and population density) and dynamic features (case number, and the information on hospitalizations across all timestamps), whereas the edges denote disease transmission. The weights of edges in the STAN are calculated in terms of geographical proximity and population size. After constructing the network structure of the STAN, the graph attention mechanism is used to update node attributes and feed them to a GRU to extract temporal features. Epidemiological constraints are incorporated into STAN learning and prediction, as it generates two types of output by multitasking prediction settings: (1) epidemiological parameter predictions (i.e., the transmission rate and recovery rate); and (2) disease dynamics predictions (i.e., the increases in the infected and recovered case numbers). Gao et al. \cite{gao2021stan} also designed the loss function for model optimization based on the above-mentioned two kinds of outputs. In their loss function, the first item is the prediction loss which captures short-term trends by calculating errors between the dynamics predicted by the deep modules and real case numbers, the second item is the epidemiological loss which captures long-term trends by calculating the errors between the disease dynamics simulated with the SIR model and real case numbers.
% The loss function is given as follows:
% \begin{equation}
% \begin{aligned}
% 	L & = \sum_i^T \sum_j^N \left( (\widehat{\Delta \boldsymbol{I}}-\Delta \boldsymbol{I})^2+(\widehat{\Delta \boldsymbol{R}}-\Delta \boldsymbol{R})^2 \right) + \left( \left(\widehat{\Delta \boldsymbol{I}}^d-\Delta \boldsymbol{I}\right)^2 + \left(\widehat{\Delta \boldsymbol{R}}^d-\Delta \boldsymbol{R}\right)^2 \right),
% \end{aligned}
% \end{equation}
% where $\Delta \boldsymbol{I}$ and $\Delta \boldsymbol{R}$ denote the ground truth of daily increased infected case number and recovered case number; $\widehat{\Delta \boldsymbol{I}}$ and $\widehat{\Delta \boldsymbol{R}}$ also denote the daily increased number of infected cases and recovered cases but are predicted by deep modules; $\widehat{\Delta \boldsymbol{I}}^d$ and $\widehat{\Delta \boldsymbol{R}}^d$ denote the daily increased number of infected cases and recovered cases which are calculated by epidemiological parameters estimated by deep modules.
The causal-based graph neural network (CausalGNN) model proposed by Wang et al. is another framework that constrains the dynamic attention-based GNN module with an epidemiological model (i.e., the SIRD model) \cite{wang2022causalgnn}. Similar to \cite{gao2021stan}, the CausalGNN model generates two types of outputs: data-driven predictions of case numbers; and the epidemiological parameters in the SIRD model, which are used to simulate predictions of case numbers. The loss function in this model consists of two $l_1$-norm items, which are errors between the ground truth of the case number and the two types of predictions. Unlike the model developed by Gao et al. \cite{gao2021stan}, the CausalGNN model also feeds the simulations obtained from the SIRD model together with the input features into the data-driven model to generate the model outputs.
%LSTM
Wang et al. \cite{wang2019defsi, wang2020tdefsi} proposed an epidemic prediction framework, named \textbf{d}eep learning based \textbf{e}pidemic \textbf{f}orecasting with \textbf{s}ynthetic \textbf{i}nformation (DEFSI), to conduct short-term and high-resolution ILI incidence prediction. The novelty of DEFSI is that it generates fine-scale ILI incidence data from an agent-based simulator (EpiFast) of an SEIR model, whose transmission parameters are estimated from the surveillance data. The obtained synthetic data are used to train a two-branch LSTM model to capture the within-season and between-season temporal dependencies of the incidence trends, and the model outputs are merged to generate final predictions.
%the epidemiological parameters (the health status of individuals, transmission rate, and incubation periods)
%\subsubsection{Multi-task learning}
%Asymmetric multi-task learning methods~\cite{nguyen2020clinical}
%~\cite{seo2020physics}

\subsubsection{Discussions: advantages and limitations}

\ 

\noindent
As aforementioned, epidemiology-inspired machine learning exploits the advantages of epidemiological models and data-driven machine learning models to construct models that are more interpretable than models built in a totally black-box manner while preserving representation and fitting capacity. Thus, epidemiology-inspired machine learning can infer disease transmission patterns from available data to generate a model that is more consistent with epidemiological constraints than black-box models. Based on the results of epidemiology-inspired machine learning models, estimated disease patterns can be further analyzed to provide more potential information hidden in data.
However, most existing epidemiology-inspired machine learning models, especially those based on neural networks, incorporate a compartmental model in a serial or parallel manner, such as by utilizing a compartmental model to generate data that are fed into the learning architecture or by predicting the parameters of epidemiological models and then calculating the output of the compartmental model as a constraint. As such, their model structures are still somewhat black-box-like and thus are insufficiently interpretable. Consequently, there is a need for the development of methods that can naturally and intrinsically integrate knowledge of epidemiological models into structures of data-driven machine learning models.

%For developing informed and precise intervention policies to keep up with the pandemic situation to cut off disease transmission and contain the disease outbreak, the following aspects are necessary:
%\begin{enumerate}
%	\item The varying dependency in terms of spatial and temporal domains. Some work obtains the transmission networks with fixed structures and values by fitting models from training data and predicts future pandemic trends based on it \textcolor{CornflowerBlue}{[add citiation]}. However, it will cause poor generalization ability when the case trends change dramatically and fast because of the domain/distributional shift problem. Some work 
%	\item 
%\end{enumerate}

\section{Challenges}
\begin{figure*}[htb!]
% \vspace{0.1cm}
  \centering
  \subfigure{\includegraphics[width=1\textwidth]{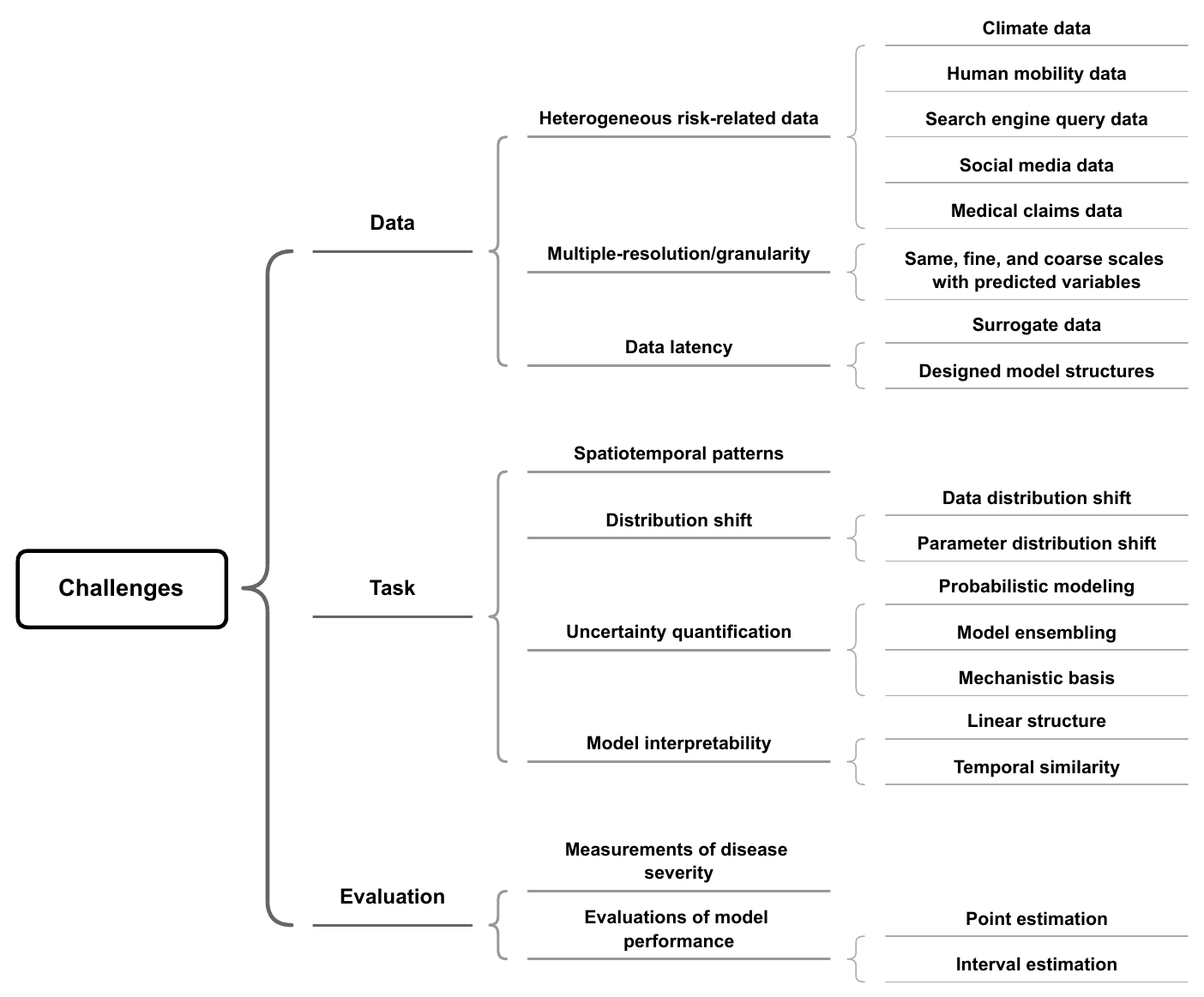}}
  \caption{Three main challenges in machine learning for infectious disease risk prediction: (1) data-related challenges (Section~\ref{Data challenges}); (2) task-related challenges (Section~\ref{Task challenges}); and (3) evaluation-related challenges (Section~\ref{Evaluation challenges}).}
  \label{fig:structure of challenges}
 % \vspace{0.1cm}
\end{figure*}

In Section 2, we introduced machine-learning methods and models that are currently used for infectious disease risk prediction and categorized the models by their structures. The plurality of model structures and their integrations that can be used to explicitly model disease transmission in different ways or directly predict disease dynamics are evident from the abundant previous literature. However, in practice, the designs of various models could benefit from specific challenges being addressed, and these have not been summarized thoroughly in previous surveys. Furthermore, there is no unique way to comprehensively classify prediction models; however, their various advantages are revealed by classifying these models in different ways. Therefore, in this section and Fig.~\ref{fig:structure of challenges}, we summarize the above-mentioned models in terms of input data, task nature, and output evaluation, to present the challenges that are met during the prediction of disease risk.

% that cover the spectrum from mechanism-based models to data-driven models

% In the previous three sections, we divide the existing works from the perspective of characteristics of model structure to illustrate the evolution from mechanism-based models to data-driven models.
%we will introduce three main challenges, i.e., data challenges, task challenges, and measurements, meeting when designing the model for epidemic prediction and demonstrate which techniques can address these challenges and how current techniques were used to address these challenges in the aforementioned works.
% re-illustrate from several different perspectives.

\subsection{Data challenges}\label{Data challenges}
The availability and limitations of disease-related data must be carefully examined to enable appropriate algorithms to be designed, i.e., algorithms that utilize or manage abundant data.

\subsubsection{Heterogeneous risk-related data}

\ 

\noindent
Because disease propagation is closely related to interactions between humans, the environment, and pathogens, data are commonly collected from the physical world. However, with the advent of the Internet and social media, non-physical types of information, such as social interactions, are increasingly reflecting transmission patterns between humans. Thus, many researchers have used abundant information from different sources by exploring and exploiting data of various risk-related factors to comprehensively characterize the patterns of disease spread. Some quantitative relationships between risk-related factors and disease dynamics have been explored and defined in empirical studies, but most of the causal links and correlations between these factors and disease dynamics are complex and non-intuitive. Therefore, when using these rich related data in prediction, many studies have either directly utilized known formulations or automatically discovered the statistical relationship between various risk-related factors to improve epidemic prediction.

\paragraph{Climate data}
One of the most significant features of the dynamics of diseases such as influenza and malaria, which are triggered or influenced by the climate, is their obvious correlation with trends in climate conditions.
For example, Jeffrey and Melvin reanalyzed data obtained in laboratory experiments on guinea pigs \cite{lowen2007influenza} and demonstrated that \textbf{a}bsolute \textbf{h}umidity (AH) greatly affected the \textbf{i}nfluenza \textbf{v}irus \textbf{t}ransmission (IVT) and \textbf{i}nfluenza \textbf{v}irus \textbf{s}urvival (IVS) in temperate regions \cite{shaman2009absolute}. On this basis, Shaman et al. further examined these findings at the human population level and proposed an AH-forced SIRS model, which quantitatively defines the relationship between AH and basic reproduction number, to simulate the seasonal patterns of influenza-related deaths \cite{shaman2010absolute}. In their subsequent studies, Shaman et al. used an AH-driven SIRS model that has a similar structure to their SIRS model \cite{shaman2010absolute} to incorporate AH data into their data-assimilation framework and generate influenza predictions for cities in the USA \cite{shaman2012forecasting, shaman2013real}.
In regions with non-temperate climate patterns, such as subtropical and tropical regions, the relationship between AH and influenza can be quite different. 
In contrast to the above-mentioned studies, which have used compartmental models to incorporate risk-related factors, Tamerius et al. collected data from numerous sites across a global range and explored the relationship between several climate variables (i.e., temperature, solar radiation, specific humidity, and precipitation) and the influenza peaks by rank order analysis. They also established univariate and multivariate logistic regression models to assess this relationship \cite{tamerius2013environmental}. Venna et al. used a symbolic time-series approach to model the nonlinear relationship between climate variables (i.e., precipitation, temperature, and sun exposure) and dynamics of influenza \cite{venna2018novel}.

% fed the  into the deep learning model to correct the XXX \cite{venna2018novel}.
% based on the compartmental model and data assimilation approaches to predict influenza

Vector-borne diseases, such as malaria, are greatly affected by climate conditions, as these influence the biological features of vectors and pathogens, such as the survival rate of mosquitoes and the incubation period of \textit{Plasmodium}. The quantitative relationship between temperature or rainfall and VCAP was defined by Ceccato et al. \cite{ceccato2012vectorial}. Shi et al. utilized temperature and rainfall data to estimate some epidemiological parameters in their model, which incorporates VCAP, to assist with the prediction of malaria dynamics which show obvious seasonal fluctuations \cite{shi2020inference}. Zhang et al. used temperature and rainfall and other disease-related data to form the feature vector of locations to learn their impact on intra-regional transmission risk with the Poisson regression model \cite{zhang2015unified}.

\paragraph{Human mobility data}
Human behavior plays an important role in infectious disease transmission \cite{funk2010modelling}, and human mobility is a crucial factor affecting the range and distribution of disease risk \cite{barbosa2018human}.
Several studies have examined the effects of different human mobility patterns on disease transmission. For instance, Balcan et al. studied two kinds of mobility patterns---short-range commuting flows and long-range flights---and explored their effects on the spatiotemporal dynamics of infectious disease by integrating a mobility network into the GLEaM model \cite{balcan2009multiscale}.
Pei et al. used commuter data from the US census survey\footnote{https://www.census.gov/topics/employment/commuting.html. Cannot be accessed on March 25, 2023} to support the prediction of influenza spread \cite{pei2018forecasting}. In their model, recurrent commuters and random visitors are considered to compose the connections between locations in the meta-population compartmental model.
Kapoor et al. utilized human mobility data from two Google mobility datasets (the Google COVID-19 Aggregated Mobility Research Dataset and Google Community Mobility Reports\footnote{https://www.google.com/covid19/mobility/. Accessed March 25, 2023}) to construct a human mobility network at the daily scale \cite{kapoor2020examining} for studying the COVID-19 pandemic. In their network, nodes represent locations, and spatial and temporal edges denote human movement between neighborhoods and connections between historical days and the current day, respectively.
Cui et al. used visitor counts in the census block group provided by SafeGraph\footnote{https://docs.safegraph.com/docs/places\#section-patterns. Accessed March 25, 2023} to model human mobility \cite{cui2021into}. Unlike the method of Kapoor et al. \cite{kapoor2020examining}, which directly constructs the edges of a human mobility network from movement data, the method of Cui et al. \cite{cui2021into} treats visitor counts and other disease-related factors as features of each location and uses the attention-based mechanism to learn the structure of a disease transmission network.

\paragraph{Search-engine query data}
In addition to physical movement, the online behavior of individuals can partly reflect the patterns of disease risk. For example, some researchers have assumed that there is a correlation between changes in ILI levels and online search activities.
For example, Ginsberg et al. developed Google Flu Trends, which uses the proportion of ILI-related search queries to overall search queries from Google as an explanatory variable for predicting ILI physician visits (the outcome) by fitting a simple linear model \cite{ginsberg2009detecting}. They focused on designing an automated method to identify the ILI-related search queries from a huge number of search records, which contribute most to the model when fitting with ILI physician-visit data. However, maintenance of the website for generating Google Flu Trends stopped in August 2015\footnote{https://ai.googleblog.com/2015/08/the-next-chapter-for-flu-trends.html. Accessed March 25, 2023}.
Nevertheless, the Google Extended Trends (GET) application programming interface remains accessible and provides the statistics of online search trends at various temporal and geographical granularities that researchers can use to train their models \cite{yang2015accurate}.
Yang et al. utilized GET data to devise the \textbf{a}uto\textbf{r}egression with \textbf{Go}ogle search data (ARGO) method for influenza epidemic estimation \cite{yang2015accurate}. They obtained the search terms that are strongly correlated with the ILI from Google Correlate\footnote{www.google.com/trends/correlate/. Cannot be accessed on March 25, 2023} (which stopped providing data after March 28, 2015) and search trends from Google Trends\footnote{https://trends.google.com/trends/. Accessed March 25, 2023}, and then used these data as input for an ARX model.
Kandula et al. also used Web-based search activity data from GET to nowcast ILI dynamics at subregional geographic scales (i.e., state-level) \cite{kandula2017subregional}. They formalized strongly correlated query terms for a specific region in a period as explanatory variables and used them to train an ARIMA model to predict the response variable (i.e., ILI observations). Then, they treated the forecast of the ARIMA model as an additional explanatory variable and used it with the original explanatory variables to train a random forest model for making final predictions of the response variable.
Similarly, search query data from Baidu\footnote{https://index.baidu.com/. Accessed March 25, 2023}, one of the biggest search engines in China, were used by Yuan et al. to monitor influenza case counts in China \cite{yuan2013monitoring}. They also used a multiple regression model to predict case counts based on the previous case count and a composite index of searches.

% And they use the predictor matrix to 
% the features which are added into the predictor matrix
% Meanwhile, they treat them as explanatory variables.
% by combing the ARIMA model and the random forest model.
% Based on the improved predictor matrix, .

\paragraph{Social media data}
Other important online behaviors of individuals that reflect the health status and contact patterns of humans are social media posts and virtual interactions. Although these data do not directly quantitatively reveal the extent and patterns of disease transmission, they can be mined to obtain some useful information.
For example, Achrekar et al. \cite{achrekar2011predicting} calculated Pearson correlation coefficients and fitted a regression model, finding a linear correlation between the number of Twitter users who tweeted influenza-related posts and the percentage of ILI physician visits. Based on this evidence, they proposed the \textbf{s}ocial \textbf{n}etwork \textbf{e}nabled \textbf{f}lu \textbf{t}rends (SNEFT) framework, which is based on an ARX model, to predict ILI case numbers from ILI physician visits and collected Twitter data.
Similarly, Zhang et al. counted ILI-related tweets to estimate the number of influenza infections in a given week and in given locations, but without filtering retweets and successive posts \cite{zhang2017forecasting}.

However, in \cite{achrekar2011predicting, zhang2017forecasting}, only the number of users who post influenza-related information or the number of tweets relevant to the influenza keywords were used to estimate the number of influenza infections; other features of textual information were not utilized. In contrast, Volkova et al. extracted detailed linguistic features and communication patterns as latent embeddings from social media posts of a Twitter dataset and fed these data, together with ILI data, into a joint neural network model based on LSTM modules to predict ILI dynamics \cite{volkova2017forecasting}. Hua et al. used a Bayesian model to identify individual health states (i.e., healthy, exposed, and infectious) from Twitter posts, aggregated these states at the population level, and then employed these states to inform the epidemiological parameters of a simulation model \cite{hua2018social}.

\paragraph{Medical claims data}
In addition to data that describe disease severity and transmission in a whole population (i.e., incidence rates or case numbers), there are abundant individual-patient data that can provide insights into disease severity and medical resource utilization.
Gao et al. extracted daily data on hospitalizations, intensive care unit stays, and frequency of diagnosis codes from IQVIA's medical claim data\footnote{https://www.iqvia.com/solutions/real-world-evidence/real-world-data-and-insights. Accessed March 25, 2023} \cite{gao2021stan} and combined these data with the number of COVID-19 cases (active, confirmed, and death cases) provided by Johns Hopkins University\footnote{https://github.com/CSSEGISandData/COVID-19. Accessed March 25, 2023} to serve as the dynamic features of nodes in a graph neural network to facilitate the characterization of spatiotemporal patterns.
In \cite{gao2022popnet}, Gao et al. extracted the statistics of disease-related features from the same IQVIA claim dataset to facilitate the training and prediction of a downstream deep learning model.
% Although epidemiology mainly studies the disease transmission patterns in the whole population instead of clinical diagnosis and treatment.

% \paragraph{Mixed heterogeneous data}
% Many works utilize heterogeneous data from multiple sources in their model, such as and

%\subsubsection{Heterogenous risk-related data}
%\begin{itemize}
%	\item Climate (Absolute Humidity~\cite{shaman2011absolute}, Temperature~\cite{tamerius2013environmental})
%	\item Search engine query data (~\cite{ginsberg2009detecting}, ~\cite{yuan2013monitoring})
%	\item Social media data (A joint neural network model~\cite{volkova2017forecasting})
%\end{itemize}

\subsubsection{Multiple-resolution/granularity}

\ 

\noindent
One of the most common problems with using data from heterogeneous sources is that data on different risk factors have different spatial and temporal resolutions. The common spatial resolutions for disease-related data, in order from coarse to fine, are region, country, province/state, county, and village, whereas temporal resolutions are year, month, week, and day.
% Mainly focus on the spatial scale?
In \cite{tan2020demystifying}, Tan et al. considered data with three types of resolution: (1) the same scale as the predicted variables (same-scale data); (2) a scale that is finer than the predicted variable (fine-scale data); (3) a scale that is coarser than the predicted variable (coarse-scale data). They designed an input module to integrate the data from heterogeneous data sources with the above-described scales as a vector and treat the vector as the input of a hierarchical RNN model. Specifically, at each time step (at the same resolution as the target variables), the fine-scale data are encoded as a vector representation by an encoder structure based on an RNN and then concatenated with the same-scale data and coarse-scale data to give an integrated vector.

% \cite{cui2021into} concatenates the case number data and the visits data with a coarser scale directly and treats them as the input of an encoder which is a component of the overall model.

%Many works align the resolution of data with the predicted outcomes' ., i.e., case number in the week or month scale. For the granularity beyond the outcome's, the ; for the granularity below the outcome's.

%Different from the above two 
%\cite{zheng2021hierst}

\subsubsection{Data latency}

\ 

\noindent
Traditional surveillance data (e.g., reports of case numbers from government institutions) on infectious diseases usually take several weeks to be published because of the processes of data collection, organization, and revision \cite{chakraborty2014forecasting, gao2022popnet}. One of the ways used to manage this challenge in recent studies of real-time prediction approaches is to use surrogate or proxy data with high timeliness \cite{chakraborty2014forecasting}, such as online search data (e.g., Google Trends\footnote{https://trends.google.com/trends/. Accessed March 25, 2023}), social media posts (e.g., Twitter) \cite{achrekar2011predicting}, news (e.g., influenza-related news stories), and meteorological data (e.g., temperature, rainfall, and humidity), to make predictions at the same time.
Another way is to use a designed model structure to deal with the latency within data. For instance, Gao et al. designed two attention mechanisms: S-LAtt and T-LAtt \cite{gao2022popnet}. In the S-LAtt module, the spatial embedding of updated data with latency is fused with the real-time data and its embedding. To obtain the embeddings of real-time data and updated data, they designed a dual GAT to learn these embeddings separately. To obtain the embedding of updated data, the attention weights are calculated by considering the spatial similarity and the degressive marginal influence of latency. In the T-LAtt module, the temporal embeddings from real-time data and updated data are fused. When the embedding of updated data is calculated, the learning of attention weights is constrained by the time latency between the real-time data and the updated data.

%\subsubsection{Data noise}
%During the data collection procedure, the data noisy is unavoidable. However, the noisy will greatly affect the model output in some case.
%%For the models without robustness, they are sensitive to the noisy.
%%Due to the collected procedure,.
%~\cite{rodriguez2021deepcovid}.
%
%\subsubsection{Data incompleteness}
%
%\subsubsection{Data sparsity}
%~\cite{rodriguez2021deepcovid}.

\subsection{Task challenges}\label{Task challenges}
In addition to the challenges associated with the heterogeneity of data, the task of infectious disease risk prediction faces many other challenges related to modeling disease transmission. These include how to capture the spatiotemporal patterns of disease transmission, how to cope with the problems of distribution shift and uncertainty, and how model structures and predictions should be interpreted.
In this section, we summarize several concerns about task-related challenges and introduce how they are addressed by various models.
% when talking about modeling disease transmission
% which are of paramount importance for achieving a good prediction goal should be taken up in earnest.
% {\color{red}The goodness of model architecture and resulting predictions} could be reflected in various senses, such as
% interpretability, prediction accuracy, robustness, etc.

\subsubsection{Characterizing spatiotemporal patterns}
%dependency/correlation

\ 

\noindent
Characterizing spatial and temporal patterns is one of the most important tasks required for determining infectious disease propagation. Herein, we discuss how temporal and spatial patterns are incorporated, respectively, into models in previous studies.

%Because of almost all XXX, we focus on distinguishing the spatial assumptions in different models in this section.

% \paragraph{{\color{red}One time step or multiple time steps ahead}}
\paragraph{Temporal modeling}
Naturally, current approaches assume that disease dynamics are related to historical dynamics, and this assumption is obviously consistent with the course of disease transmission.
Some models assume that the dynamics at the current time step develop from the dynamics at the immediately previous time step, such as the ODEs of the SIR model and other similar compartmental models \cite{kermack1927contribution, hethcote1991some}, or that they are related to the dynamics of multiple time steps ahead, due to the existence of disease incubation, such as autoregressive-type models \cite{dettling2013applied, lutkepohl2004applied}, and many deep learning models based on RNN modules \cite{volkova2017forecasting, venna2018novel, wu2018deep, zheng2021hierst, kamarthi2021doubt, rodriguez2021steering, gao2022popnet}, CNN modules \cite{deng2020cola, cui2021into}, or GNN modules \cite{cao2020spectral}.
Specifically, unlike traditional RNN modules, which can naturally consider multiple time steps, some CNN-based deep learning models capture the temporal patterns at different time scales by conducting dilated convolution across time \cite{deng2020cola} or encode temporal patterns in various time windows by using kernels of different sizes \cite{cui2021into}. There is also a GNN-based model \cite{cao2020spectral} that uses the GFT to capture the inter-series temporal dependency in the spectral domain, and uses the DFT to learn the intra-series in the frequency domain.

% spectral graph convolution

% \paragraph{{\color{red}Single location or multiple locations}}
\paragraph{Spatial modeling}
% With the similar sense
Similarly, the spatial pattern is also a necessary characteristic in the description of disease propagation. Some studies have focused on the modeling of disease transmission in a specific location, such as compartmental models at the population level \cite{kermack1927contribution, hethcote1991some} and machine learning models for single locations \cite{volkova2017forecasting, venna2018novel, zimmer2020influenza}, whereas many other studies have predicted disease dynamics at multiples locations simultaneously. Some of these studies have assumed that spatial patterns can be reflected by the statistical correlation of data between different locations; that is, locations with similar feature conditions have similar disease severities. For instance, the GP regression model \cite{senanayake2016predicting} uses designed kernels to calculate the covariance matrix, which reflects spatial similarity.
%Some works do not explicitly infer the network structure by the spatial module but directly use the deep learning models to make the prediction of multiple locations \cite{tan2020demystifying}.
Other studies have modeled disease transmission between locations by assuming that a disease transmission network exists, such as compartmental models at the meta-population level \cite{pei2020initial} and the individual level \cite{hoertel2020stochastic}, a multivariate regression model \cite{zhang2015unified, pei2018group}, and deep learning models based on a CNN module \cite{wu2018deep} or a GNN module \cite{kapoor2020examining, deng2020cola, zheng2021hierst, gao2022popnet}.
The nodes of disease transmission networks typically consist of locations or individuals of interest, whereas edges consist of the interactions or relationships between locations or individuals. Although network-based methods aim to achieve the same goal---i.e., generate epidemic predictions for multiple locations by considering their relationships---they have different focuses on adopted network hypotheses, based on which they can be distinguished. However, although such methods assume that a disease transmission network exists, the interactions in a transmission network cannot be observed directly in the real world, which means there is no ground truth for a disease transmission network. Therefore, some ODE-based models construct a transmission network based on physical contact patterns, and some machine learning models infer network structure from data during model optimization. In doing so, various hypotheses are made regarding network properties to approximate the real situation. In the following, we introduce three types of network structure hypotheses that have often been used in previous studies.

%Properties of network structures
% Undirected or directed networks
The first common type of setting of network properties is whether edges in a network are undirected or directed, where an undirected edge and a directed edge denote a symmetrical and an asymmetrical correlation or dependency between different locations, respectively. In studies that have assumed that a network is undirected (e.g., \cite{gao2022popnet}), some symmetrical similarity measurements of features have been considered as the quantitative relations between different locations. This implies that the disease dynamic of locations with similar conditions can be used to improve the prediction in a target location.
In other studies, a network has been assumed to be directed \cite{zhang2015unified, pei2018group, wu2018deep, kapoor2020examining, deng2020cola, zheng2021hierst} which implies that there are asymmetrical interactions between different locations. Some of these studies have incorporated prior knowledge and data on the drivers of disease transmission, such as human mobility, which causes cross-regional disease transmission, to assist with network inference\cite{zhang2015unified, kapoor2020examining}. Other studies have directly inferred a transmission network from disease dynamic data \cite{pei2018group, wu2018deep, pei2020active, deng2020cola, zheng2021hierst}.
% \cite{kapoor2020examining} uses the daily human mobility data to construct network structures; \cite{deng2020cola} uses the attention-mechanism to learn the dynamic networks; ; \cite{zheng2021hierst}.

% Static or dynamic networks
The second common type of setting of network properties is whether the network structure is static or dynamic. Some studies have used a fixed network to predict disease risk~\cite{wu2018deep, pei2018group, zhang2015unified}, which means that after a model is trained on historical data, an inferred transmission network remains constant during the whole prediction.
% which means that network structure is learned from the historical training data and will 
However, in the real world, the structure of a disease transmission network changes over time, due to changes in risk-related factors (e.g., climate, socio-economic indicators, and human behaviors) and interactions between these factors. Therefore, many studies have tried to characterize dynamic disease transmission networks \cite{kapoor2020examining, liu2023assessing}.
% {\color{pink}(add some citations and more descriptions)}.
%treat the development of disease dynamics as the transmission process in a network,
%or individual level \cite{hoertel2020stochastic} and
% implies there are causal links
% from location $A$ to location $B$ will impact the disease propagation in location $B$
% bring the imported disease cases
% (current)

% Advanced network properties
In addition to basic network properties, more advanced network characteristics such as the power law of degree distribution are ubiquitous in real-world networks, such as the World Wide Web, worldwide airline networks, and interurban commuting networks \cite{barrat2008dynamical, barthelemy2011spatial, barabasi2013network}.
Some studies have used this power law to constrain network structure learning for disease transmission. Specifically, Wang et al. \cite{wang2018inferring} adopted a metapopulation SIR model to model disease propagation between different locations and thereby construct an infection network. That is, by assuming that the node degrees of the disease transmission network followed the power-law distribution, they formalized the prior probability of the whole network and incorporated it into their Bayesian framework of network inference.
However, in recent studies that have inferred disease transmission networks, this aspect has not been fully explored.

% disease transmission networks

%\begin{table*}[h]
%\begin{center}
%\begin{minipage}{\textwidth}
%\caption{Summary of work}\label{tabUncertainty}
%\begin{tabular*}{\textwidth}{@{\extracolsep{\fill}}ccc@{\extracolsep{\fill}}}
%\toprule%
%%& \multicolumn{3}{@{}c@{}}{
%%Element 1\footnotemark[1]} & \multicolumn{3}{@{}c@{}}{Element 2\footnotemark[2]} \\\cmidrule{2-4}\cmidrule{5-7}%
%Patterns & Types & Components \\
%\midrule
%%CGP~\cite{qian2020and} & Gaussian process (GP) & \\
%Temporal pattern & One time step & \\
% & multiple time steps & \\
%Spatial pattern & Single location & \\
% & multiple locations & \\
%\botrule%
%%\makecell[c]{Calibration Score (CS) \\ Calibration Plot (CP) \\ Defined by the paper}
%\end{tabular*}
%\end{minipage}
%\end{center}
%\end{table*}

\subsubsection{Distribution shift}

\ 

\noindent
Generally, machine learning methods that are trained based on empirical risk minimization face an inherent issue: the generalization ability. A model requires good ability of generalization to make accurate predictions when receiving inputs that it has never seen.
However, machine learning models for epidemic prediction struggle to achieve good generalization ability. Moreover, as epidemic trends can change quickly in a short period due to complex interactions between multiple factors, such as intervention strategies and climate conditions, the problem of distribution shift arises \cite{amodei2016concrete, kouw2018introduction}.
A few studies have examined distribution shifts as part of the topic of epidemic prediction. Wang et al. \cite{wang2021bridging} investigated two distribution shift scenarios: data distribution shift and parameter distribution shift. For each scenario, they studied interpolation and extrapolation tasks via machine learning. The extrapolation task can be regarded as model learning with distribution shift, which means that the distribution of the data or system parameters that need to be predicted is different from the distribution of the data or system parameters that are used for model training. The interpolation task is associated with a situation without distribution shift, which most current machine learning models can handle well. Wang et al. \cite{wang2021bridging} showed that physics-based mechanistic models outperform deep learning models in both of the above-mentioned scenarios, which suggests that it is possible to improve the generalization ability of deep learning models by introducing the inductive bias of mechanism-based models.
% Some models address this problem by introducing inductive bias. This kind of inductive bias could come from the specific domain knowledge of disease transmission which is discovered by empirical study. For instance, \cite{wang2021bridging}
%then causes poor generalization of constructed models in practical applications, especially deep learning models~\cite{wang2021bridging}.
%Therefore, when targeting to predict future disease trends,

\subsubsection{Uncertainty quantification}

\ 

\noindent
% I think the current classfication is good, just like the Bayesian and Frequentist is statistics; but we need to clarify the concept of accuracy/uncertainty or point/interval estimation, maybe we can avoid the word "accuracy" here.
% can be measured fair well by existing measurements
As epidemic predictions are closely related to the development and establishment of public-health intervention strategies, predictions must be both accurate and reliable to enable decision-makers to make good decisions. Usually, point estimation is used to represent a model's output and assess the model's accuracy. However, although calculating the errors between point estimations and observed data is a good way to determine a model's performance, it is insufficient to enable the development of good intervention strategies. That is, when applying epidemic prediction models in the real world, flawed data, incomplete understanding of disease transmission, unknown future potential changes, and even model design bring significant uncertainty into the results and model parameterization \cite{holmdahl2020wrong}. Therefore, many studies have generated interval estimates, which provide not only estimated values but also their confidence intervals for model outputs or model parameters.

Generally, uncertainty quantification methods can be classified into two categories: intrinsic and extrinsic methods \cite{shen2022post}. Intrinsic methods generate predictions and uncertainty estimates simultaneously. Extrinsic methods train auxiliary or meta-models to give confidence estimates in a post-hoc manner.
Current models for epidemic prediction tend to generate uncertainty measurements in an intrinsic way. Based on the ways in which they incorporate uncertainty, models can be divided into two classes: Bayesian learning-based models and ensembling models.
% stochastic processes-based models
% \cite{kamarthi2021doubt}
% how to appropriately assess prediction uncertainty is also a crucial problem.
% There are several ways to incorporate uncertainty into models and assess the uncertainty of results. We adopted the classification similar in \cite{kamarthi2021doubt} which divides existing work into three classes which are based on the Bayesian framework, stochastic processes, and model ensembling respectively.

% \paragraph{Bayesian learning}
\paragraph{Probabilistic modeling}
% Adaptively weighted ensembles probabilistic modeling \cite{brooks2018nonmechanistic}.
% disease-related observations
The first category of model is based on probabilistic modeling, which takes uncertainty into account by incorporating the probabilistic distribution of parameters or functions. Usually, these models first assign a prior distribution for targeted variables, and then use Bayes' theorem to calculate its posterior distribution.
This category can be subdivided into two classes. The first class assumes that model parameters follow a probabilistic distribution. For instance, the empirical Bayes framework proposed by Brooks et al. to predict ILI trends uses historical data to estimate the prior distribution of model parameters and produces the posterior distribution of epidemic curves \cite{brooks2015flexible}. Other Bayesian inference methods, including the Kalman filtering method and its variants \cite{anderson2001ensemble}, and PFs \cite{arulampalam2002tutorial}, have also been used in disease dynamic prediction \cite{shaman2012forecasting, tizzoni2012real, shaman2013real, yang2015inference, zhang2017forecasting, pei2018forecasting}.
% data assimilation
% Data assimilation technique \cite{shaman2012forecasting}.
% Global Epidemic and Mobility Model \cite{tizzoni2012real}.
% a data-driven spatial, stochastic, and individual-based epidemic model (global epidemic and mobility model \cite{zhang2017forecasting}.
% An ensemble forecast system for predicting the spatiotemporal spread of influenza
% \cite{pei2018forecasting}.
Rather than assuming that the model parameters follow a probability distribution, stochastic process-based models (e.g., the GP model) define the probability distribution over functions \cite{bishop2006pattern}. Some representative studies have used the GP model to predict disease dynamics and provide the uncertainty of results.
For instance, Senanayake et al. used the Gaussian process regression model with designed spatiotemporal kernels to model the spatiotemporal characteristics of influenza transmission and used variational inference to optimize the model to adapt to a large dataset \cite{senanayake2016predicting};
Zimmer and Yaesoubi trained an independent GP model for each location to capture the correlation between historical data in the previous season and predict the dynamics $t$ weeks ahead based on available data for the current year \cite{zimmer2020influenza}.
% Stochastic processes and DNNs
Furthermore, some studies have used deep learning models to perform stochastic processes; this type of model is called \textbf{n}eural \textbf{p}rocesses (NPs) \cite{garnelo2018neural}. Its extensions, such as the \textbf{f}unctional \textbf{n}eural \textbf{p}rocess (FNP) \cite{louizos2019functional} and the \textbf{r}ecurrent \textbf{n}eural \textbf{p}rocess (RNP) \cite{qin2019recurrent, kim2019attentive}, have also been developed to capture complex dependency.
For instance, the EPIFNP is a neural process model that incorporates RNP and FNP modules for disease dynamics prediction \cite{kamarthi2021doubt}. It uses a probabilistic deep-sequence model to encode the sequence data of influenza outbreaks as latent variables, and uses a stochastic correlation graph to learn the correlations between these variables. Finally, it uses an MLP module to parameterize the predictive distribution of model outcomes.

\paragraph{Model ensembling}
Another way is to use ensemble approaches, which collect a set of predicted values from multiple trained models and calculate their probabilistic distributions to enhance the robustness of predictions.
For instance, the COVID-19 Forecast Hub \footnote{https://covid19forecasthub.org. Accessed March 25, 2023} ensembles the forecasts from different models from various institutions to generate US COVID-19 death data \cite{ray2020ensemble}; and
\cite{rodriguez2021deepcovid} evaluates the uncertainty from data by using bootstrapping, which is a data sampling technique, to re-sample an entire training dataset to multiple subsets and then uses the subsets to train a set of models. Based on the trained models, multiple prediction results are produced and can be used to calculate the confidence interval.

%is critical for the decision-making of public policy related to containing infectious diseases.  the  of public health 
% \cite{guo2017calibration, gal2016dropout}.

% \begin{table*}[h]
% \begin{center}
% \begin{minipage}{\textwidth}
% \caption{Summary of work including uncertainty quantification.}
% \label{tab:Uncertainty}
% \begin{tabular*}{\textwidth}{@{\extracolsep{\fill}}ccc@{\extracolsep{\fill}}}
% \toprule%
% %& \multicolumn{3}{@{}c@{}}{
% %Element 1\footnotemark[1]} & \multicolumn{3}{@{}c@{}}{Element 2\footnotemark[2]} \\\cmidrule{2-4}\cmidrule{5-7}%
% Work & Components & Calibration metrics \\
% \midrule
% %CGP~\cite{qian2020and} & Gaussian process (GP) & \\
% EPIFNP \cite{kamarthi2021doubt} & Functional Neural Process (FNP) & \makecell[c]{Calibration Score (CS) \\ Calibration Plot (CP) \\ Defined by the paper} \\
% \cite{senanayake2016predicting} & Variational Gaussian Process Regression & \\
% \cite{brooks2018nonmechanistic} & Delta densities & \\
% EB \cite{brooks2015flexible} & Semiparametric Empirical Bayes framework & \\
% \cite{zimmer2020influenza} & Gaussian Processes & log-score \\
% \botrule
% \end{tabular*}

% %\tabincell

% %\footnotetext{Note: This is an example of table footnote. This is an example of table footnote this is an example of table footnote this is an example of~table footnote this is an example of table footnote.}
% %\footnotetext[1]{Example for a first table footnote.}
% %\footnotetext[2]{Example for a second table footnote.}
% \end{minipage}
% \end{center}
% \end{table*}

\subsubsection{Model interpretability}

\ 

\noindent
In addition to the broad practical applications of deep learning models in other domains closely related to human well-being, such as healthcare \cite{ahmad2018interpretable}, the interpretability of deep learning models has been examined, and its rigorous definition and associated challenges have been widely discussed \cite{doshi2017towards, rudin2022interpretable}. Moreover, the interpretability of epidemic prediction models is pivotal, because inappropriate interpretation of models and incomplete understanding of results may lead to unsound decisions, which could adversely affect human well-being and waste precious anti-epidemic resources. To avoid this kind of loss as much possible, researchers need to be cautious when they interpret models and obtain results.

Two widely known and intuitive interpretable models for disease transmission are the classic mechanism-based models (e.g., the SIR model \cite{kermack1927contribution} and the SEIR model \cite{hethcote1991some, li1995global}), which explicitly describe and simulate the disease transmission process by using ODEs in which the relationships between various variables are mathematically defined. 

Recently, some machine learning models have also explored interpretability. In particular, some models incorporate machine learning methods to infer the epidemiological parameters of compartmental models \cite{arik2020interpretable}, whereas other models use a linear model structure, such as AR and MA-based models \cite{dettling2013applied, lutkepohl2004applied}, which assume that a prediction is the weighted sum of historical dynamics.
Deep learning models are usually treated as black-box models because the relationships between input and output are highly non-linear and are implicitly encoded by the model structure and learned parameters. However, recently, many researchers have explored the possibility of incorporating explainable elements into deep learning model structures. Thus, some studies have used the similarity of time series to explain predictions. For example, \cite{adhikari2019epideep} assumed that the current (to be predicted) season is similar to some historical seasons and that this similarity can be used to aid the prediction of the incidence curve of the current season. Based on this assumption, the deep learning modules are first used to learn the similarity between historical trends by clustering and then the incomplete data of the current season are mapped to the closest historical season in the latent space. The approach in \cite{kamarthi2021doubt} is based on similar assumptions and uses a functional neural process module to learn the correlation between the predicted season and past seasons.
% \cite{rodriguez2021deepcovid} develops data ablation and data visualization modules with a web-based interface for user interaction to provide information about the impact of the data signals on predictions and the historical trends of data signals.

% \subsubsection{Prediction type}
% There are also different types of prediction.

% \paragraph{Nowcasting}
% Nowcasting, also known as real-time prediction, is predicting the current situation. In doing this type of prediction, at the time slot which tends to predict, the outcome of interest is usually unknown, but covariates can be obtained from other sources. Therefore, the predictions of the outcome can be based on the data of the outcome and covariates in historical duration and the covariates in the prediction slot.

% \paragraph{Forecasting}
% Forecasting is about predicting future dynamics without any other related information during the time slot of prediction. In this situation, the prediction is only based on historical data because the .

\subsection{Evaluation challenges}\label{Evaluation challenges}
When constructing models and evaluating their performance, many different types of outcomes and evaluation measurements can be involved, depending on data availability and practical needs. How to identify and use appropriate measurements is thus a challenge to the modeling task. In the following section, we summarize the outcomes of prediction models and measurements that have been widely used to evaluate these outcomes.

\subsubsection{Measurements of disease severity}

\ 

\noindent
When constructing an epidemic prediction model, one of the most important tasks is to determine the model outcome, which is usually a measurement of disease severity in the target population. In general, the choice of predicted variables is based on the goals of public health policy and on data availability. Various indicators of disease severity have been used. The most commonly used indicators include disease incidence \cite{wang2019defsi, rodriguez2021deepcovid}, case numbers \cite{pei2018group, zhang2015unified, senanayake2016predicting, kapoor2020examining, cui2021into, deng2020cola, wu2018deep, zheng2021hierst}, death counts \cite{kapoor2020examining, cui2021into, zheng2021hierst}, patient visit counts related to the disease \cite{deng2020cola, kamarthi2021doubt}, and disease activity levels \cite{deng2020cola, wu2018deep}.
In addition, some specialized indicators have been used to describe the seasonal outbreak of influenza, such as peak intensity, peak time, final epidemic size, onset time, and duration of outbreaks \cite{zhang2017forecasting, zimmer2020influenza, brooks2015flexible}.

\subsubsection{Evaluation of model performance}

\ 

\noindent
The above introduction to previous studies shows that some models generate point estimations that can be directly compared with observed data. Some researchers have considered uncertainty in their model designs, and have therefore presented their predictions in interval/quantile-based format due to the requirements of practical use (e.g., \cite{bracher2021evaluating}). This accounts for the different evaluation methods that have been used for these two kinds of output formats: point estimation and interval estimation.
% The outcome of prediction models includes , so there are different ways adopted to make corresponding evaluations.

\paragraph{Point estimation}
The most common methods used to evaluate the accuracy of point estimation include the \textbf{r}oot-\textbf{m}ean-\textbf{s}quare \textbf{e}rror (RMSE), as shown in Eq.(\ref{Eq:RMSE}); the \textbf{m}ean \textbf{a}bsolute \textbf{e}rror (MAE), as shown in Eq.(\ref{Eq:MAE}); the \textbf{m}ean \textbf{a}bsolute \textbf{p}ercentage \textbf{e}rror (MAPE), as shown in Eq.(\ref{Eq:MAPE}); and the \textbf{r}oot \textbf{m}ean \textbf{s}quared \textbf{p}ercent \textbf{e}rror (RMSPE), as shown in Eq.(\ref{Eq:RMSPE}). These indicators all calculate the deviation of predicted values $Y^*$ from ground truth $Y$, where $Y_t^*$ and $Y_t$ denote the predicted value and the ground truth at time step $t$, respectively.
The Pearson \textbf{corr}elation coefficient (CORR), as shown in Eq.(\ref{Eq:CORR}), is used to evaluate the correlation between predicted trend $Y^*$ and real trend $Y$; where $\bar{Y^{*}}$ and $\bar{Y}$ denote the mean value of the predicted trend and the real trend, respectively, during the time slot from $1$ to $T$.
% {\color{pink}(add citations)}

\begin{equation}\label{Eq:RMSE}
	RMSE = \sqrt{\frac{1}{T} \sum_{t=1}^T\left(Y_{t}-Y^*_{t}\right)^2}
\end{equation}
\begin{equation}\label{Eq:MAE}
	MAE= \frac{1}{T}\sum_{t=1}^{T}\vert Y_t - Y^*_t\vert
\end{equation}
\begin{equation}\label{Eq:MAPE}
	MAPE= \left(\max_{t} \frac{\vert Y_{t}-Y^*_{t}\vert}{Y_{t}}\right)*100
\end{equation}
\begin{equation}\label{Eq:RMSPE}
	RMSPE= \sqrt{\frac{1}{T} \sum^{T}_{t=1}\left(\frac{Y_{t}-Y^*_{t}}{Y_{t}}\right)^2}
\end{equation}
\begin{equation}\label{Eq:CORR}
	CORR=\frac{\left.\sum_{t=1}^T\left(Y_{t}-\bar{Y}\right)\left(Y^*_{t}-\bar{Y^{*}}\right)\right)}{\sqrt{\sum_{t=1}^T\left(Y_{t}-\bar{Y}\right)^2} \sqrt{\sum_{t=1}^T\left(Y^{*}_{t}-\bar{Y^{*}}\right)^2}}
\end{equation}

\paragraph{Interval estimation}
Some indicators are widely used for interval estimation \cite{bracher2021evaluating}. For instance, prediction interval coverage, denoted as $k_M(c)$, calculates the percentage of observed values falling into the $c$ (i.e., 50\% or 95\%) confidence interval of predicted distributions of $M$ \cite{ray2020ensemble}; \textbf{c}alibration \textbf{s}core is the integral of $\|k_M(c)-c\|$ over $c$ from $0,\cdots,1$, as shown in Eq.(\ref{Eq:CS}) \cite{kamarthi2021doubt}, and a calibration plot shows the relationship between $c$ and $k_M(c)$ \cite{kamarthi2021doubt}.
Another indicator is the logarithmic score, also called the log-score, which is used in the CDC's influenza prediction Challenge in the US. It is calculated as follows: given the predicted distribution of the outcome, first calculate the sum of probability of bins within a given interval around the true value, and then determine the natural logarithm of the calculated sum, which represents the final score \cite{zimmer2018use, zimmer2020influenza}.
\begin{equation}\label{Eq:CS}
\begin{aligned}
CS(M) & =\int_{0}^{1}\|k_M(c)-c\| \\
    & \approx 0.01 \sum_{c\in\{0, 0.01, \cdots, 1\}}\|k_M(c)-c\|
\end{aligned}
\end{equation}

\section{Conclusions and future directions}
In this paper, we review the development of machine learning in the realm of infectious disease risk prediction. First, we introduce previous studies that have explored three types of approaches to risk prediction: statistical prediction, data-driven machine learning, and epidemiology-inspired machine learning. In each category, we depict the relationship and differences between different approaches by subdividing them according to their structural characteristics. In addition, we summarize common challenges encountered when dealing with inputs, designing prediction algorithms, and conducting performance evaluation, and we also discuss several related challenges and provide examples of how previous studies have coped with these challenges.
% mechanism-based models, data-driven models, and hybrid models.
% Through the summary of all three types of models, we try to comb the transition from mechanism-based models and data-driven models and their fusion, aka hybrid models.

In research on the fusion of epidemiological models with data-driven models to aid in disease risk prediction, some directions that have been exploited are worthy of further exploration. With the increasing use of approaches that employ sophisticated deep learning structures to infer the spatiotemporal patterns of disease transmission, the problems of model overfitting and poor generalization are highlighted. Many studies of epidemiology-inspired machine learning have attempted to combine disease transmission mechanisms based on domain knowledge with data-driven models to guide the structure construction of models, regularize model parameters, and constrain model learning, and thereby generate rational model structures and meaningful results.

Specifically, epidemiological parameterization and epidemiology-embedded learning both endow models with the power of expert knowledge and data but in different ways. Currently, most models primarily comprise a white-box structure (such as epidemiological models) or a black-box structure (such as deep learning models), with the corresponding black- or white-box structure playing a subsidiary role. Therefore, it remains to be explored how to more naturally and intrinsically merge these two types of structures to enhance the interpretability of models and results.

Another important point is that the mechanism that guides data-driven machine learning models is not and should not be limited to domain knowledge of the biological process in disease transmission but should rather be based on prior knowledge from multidisciplinary fields. For example, when modeling a disease transmission network, network science theory can be applied to constrain model learning to force the learned patterns to follow physical laws. Similarly, in learning settings, the constraints of network properties for a specific network can be linked to the network properties that are empirically validated by real networks. However, there has been very little work in this direction. Therefore, improving the learning and optimization of disease transmission networks by incorporating more network properties than have previously been incorporated could be a promising future direction. Furthermore, because disease transmission networks are unobservable, analyzing the traits of inferred network structures with tools from the field of complex network research to verify these structures' rationality and provide feedback into the model could be another promising future direction.

% The network structures should be inferred reasonably and effectively
% We found that many works of hybrid models tend to improve the results' interpretability by combining the ;

%\bmhead{Supplementary information}
%Supplementary information.

% \section*{Acknowledgments}
% Acknowledgments are not compulsory. 
% Professional English language editing support provided by AsiaEdit (asiaedit.com).

\bibliography{cite}
\bibliographystyle{unsrt}

\end{document}